\documentclass[11pt,a4paper]{article}
\usepackage[hyperref]{emnlp2020}
\hypersetup{
pdftitle={Do sequence-to-sequence VAEs learn global features of sentences?},
pdfauthor={Tom Bosc, Pascal Vincent},
}
\usepackage{times}
\usepackage{latexsym}

\usepackage{multirow}
\usepackage{multicol}
\usepackage{adjustbox}
\usepackage{amsmath}

\usepackage{array}
\newcolumntype{H}{>{\setbox0=\hbox\bgroup}c<{\egroup}@{}}

\usepackage{amsmath,amsfonts,bm}

\def\eqref#1{equation~\ref{#1}}

\def\1{\bm{1}}

\def\eps{{\epsilon}}

\DeclareMathAlphabet{\mathsfit}{\encodingdefault}{\sfdefault}{m}{sl}
\SetMathAlphabet{\mathsfit}{bold}{\encodingdefault}{\sfdefault}{bx}{n}

\newcommand{\E}{\mathbb{E}}

\newcommand{\softmax}{\mathrm{softmax}}

\newcommand{\KL}{D_{\mathrm{KL}}}
\newcommand{\Var}{\mathrm{Var}}

\DeclareMathOperator*{\argmax}{arg\,max}

\usepackage{natbib}
\usepackage{tabularx,ragged2e} 
\newcolumntype{L}{>{\arraybackslash}X} 
\usepackage{microtype}
\usepackage{graphicx}
\usepackage{subfigure}
\usepackage{booktabs} 

\usepackage{microtype}

\aclfinalcopy 
\def\aclpaperid{2430} 

\title{Do sequence-to-sequence VAEs learn global features of sentences?}

\author{Tom Bosc \\
  Mila, Universit\'e de Montr\'eal \\
  \texttt{bosct@mila.quebec} \\\And
  Pascal Vincent \\
  Mila, Universit\'e de Montr\'eal, CIFAR \\
  \texttt{vincentp@iro.umontreal.ca} \\}

\date{}

\begin{document}
\maketitle
\begin{abstract}

Autoregressive language models are powerful and relatively easy to train. However, these models are usually trained without explicit conditioning labels and do not offer easy ways to control global aspects such as sentiment or topic during generation. \citet{bowman_generating_2016} adapted the Variational Autoencoder (\textit{VAE}) for natural language with the sequence-to-sequence architecture and claimed that the latent vector was able to capture such global features in an unsupervised manner. 
We question this claim. We measure which words benefit most from the latent information by decomposing the reconstruction loss per position in the sentence. Using this method, we find that VAEs are prone to memorizing the first words and the sentence length, producing local features of limited usefulness. To alleviate this, we investigate alternative architectures based on bag-of-words assumptions and language model pretraining. These variants learn latent variables that are more global, i.e., more predictive of topic or sentiment labels. Moreover, using reconstructions, we observe that they decrease memorization: the first word and the sentence length are not recovered as accurately than with the baselines, consequently yielding more diverse reconstructions.
\end{abstract}

\section{Introduction}

The problem of generating natural language underlies many classical NLP tasks such as translation, summarization, paraphrasing, etc. The problem is often formulated as learning a probabilistic model of sentences, then searching for probable sentences under this model. Expressive language models are typically built using neural networks \citep{bengio2003neural,mikolov2010recurrent}.

Whether based on LSTMs \citep{hochreiter_long_1997,sundermeyer2012lstm} or Transformers \citep{vaswani_attention_2017,radford_language_2019}, language models are mostly autoregressive: the probability of a sentence is the product of the probability of each word given the previous words. By contrast, \citet{bowman_generating_2016} built a Variational Autoencoder (\textit{VAE}) \citep{kingma_auto-encoding_2013,rezende_stochastic_2014} out of a sequence-to-sequence architecture (\textit{seq2seq}) \citep{sutskever_sequence_2014}. It generates text in a two-step process: first, a latent vector is sampled from a prior distribution; then, words are sampled from the probability distribution produced by the autoregressive decoder, conditionally on the latent vector. The goal was to encourage a useful information decomposition, where latent vectors would ``explicitly model holistic properties of sentences such as style, topic, and high-level syntactic features'' \citep{bowman_generating_2016}, while the more local correlations would be handled by the recurrent decoder.

In principle, such a decomposition can be the basis for many applications. For example, using a single, unannotated corpus, it could enable paraphrasing \citep{roy2019unsupervised} or style transfer \citep{xu2019unsupervised}. For tasks requiring conditional generation such as machine translation or dialogue modeling, we could enforce a level of formality or impose a certain tone by clamping the latent vector. Moreover, latent-variable models can represent multimodal distributions. Thus, for these conditional tasks, the latent variable can be used as a source of stochasticity to ensure more diverse translations \citep{pagnoni2018conditional} or answers in a dialogue \citep{serban2017hierarchical}.

Despite its conceptual appeal, \citet{bowman_generating_2016}'s VAE suffers from the \textit{posterior collapse} problem:
early on during training, the KL term in the VAE optimization objective goes to 0, such that the approximate posterior becomes the prior and no information is encoded in the latent variable. 
Free bits are a popular workaround \citep{kingma_improved_2016} to ensure that the KL term is above a certain level, thereby enforcing that \textit{some} information about the input is encoded. But this information is not necessarily global. After all, posterior collapse can be solved trivially, without any learning, using encoders that copy parts of the inputs in the latent variable, yielding very local and useless features.

In Section \ref{diagnosis}, we show that encoders learn to partially memorize the first few words and the document lengths, as was first discovered by \citet{kim_semi_2018}. To do so, we compare the average values of the reconstruction loss at different positions in the sentence to that of an unconditional language model. We elaborate on the negative consequences of this finding for generative models of texts.
In Section \ref{variants}, we propose three simple variants of the model and the training procedure, in order to alleviate memorization and to yield more useful global features. 
In Section \ref{section_ssl}, we empirically confirm that our variants produce more global features, i.e., features more predictive of global aspects of documents such as topic and sentiment. 
They do so while memorizing the first word and the sentence length less often, as shown in Section \ref{section_text_gen}.

\section{Model and datasets}

Firstly, we describe the VAE based on the seq2seq architecture of \citet{bowman_generating_2016}. A document, sentence or paragraph, of $L$ words $x=(x_1,\ldots,x_L)$ is embedded in $L$ vectors $(e_1,\ldots,e_L)$. An LSTM encoder processes these embeddings to produce hidden states:
$$
h_1,\ldots,h_L = \mathrm{LSTM}(e_1,\ldots,e_L)
$$
In general, the encoder produces a vector $r$ that represents the entire document. In the original model, this vector is the hidden state of the last word $r = h_L$, but we introduce variants later on. This representation is transformed by linear functions $L_1$ and $L_2$, yielding the variational parameters that are specific to each input document:
\begin{align*}
\mu &= L_1 r \\
\sigma^2 &= \exp(L_2 r)
\end{align*}
These two vectors of dimension $d$ fully determine the approximate posterior, a multivariate normal with a diagonal covariance matrix, $q_{\phi}(z|x) = \mathcal{N}(z|\mu, \mathrm{diag}(\sigma^2))$, where $\phi$ is the set of all encoder parameters (the parameters of the LSTM, $L_1$ and $L_2$). Then, a sample $z$ is drawn from the approximate posterior, and the decoder, another LSTM, produces a sequence of hidden states:
$$
h'_1,\ldots,h'_L = \mathrm{LSTM}([e_{\mathrm{BOS}};z],[e_1;z],\ldots,e_L;z])
$$
where $\mathrm{BOS}$ is a special token indicating the beginning of the sentence and $[\cdot;\cdot]$ denotes the concatenation of vectors. Finally, each hidden state at position $i$ is transformed to produce a probability distribution of the word at position $i+1$:
$$
    p_{\theta}(x_{i+1}|x_{1,\ldots,i},z) = \mathrm{softmax}(W h'_{i} + b)
$$
where $\mathrm{softmax}(v_i) = e^{v_i} / \sum_j e^{v_j}$ and $\theta$ is the set of parameters of the decoder (the parameters of the LSTM decoder, $W$ and $b$). An EOS token indicating the end of the sentence is appended to every document.

For each document $x$, the lower-bound on the marginal log-likelihood (\textit{ELBo}) is:
\begin{align*}
         \mathrm{ELBo}(x, \phi, \theta) & = - \KL(q_{\phi}(z|x)||p(z)) + \\
		 & ~~~~~~~ \E_{q_\phi}[\log p_{\theta}(x|z)] \\
		& \leq \log p(x)
        \label{eq:elbo}
\end{align*}

On the entire training set $\{x^{(1)},.,x^{(N)}\}$, the objective is:
$$
          \argmax_{\phi, \theta} \sum_{j=1}^N \mathrm{ELBo}(x^{(j)}, \phi, \theta)
$$

\subsection{Dealing with posterior collapse}

Following \citet{alemi_fixing_2018}, we call the average value of the KL term the \textit{rate}. It measures how much information is encoded on average about the datapoint $x$ by the approximate posterior $q_{\phi}(z|x)$. When the rate goes to 0, the posterior is said to \textit{collapse}, meaning that $q_{\phi}(z|x) \approx p(z)$ and that the latent variable $z$ sampled to train the decoder does not contain any information about the input $x$.

To prevent this, we can modify the KL term to make sure it is above a target rate using a variety of techniques (see Appendix \ref{subsection_freebits} for a small survey). We use the free bits formulation of the $\delta$-VAE  \citep{razavi_preventing_2019}. For a desired rate $\lambda$, the modified negative ELBo is:
\begin{align*}
\max(\KL(q_{\phi}(z|x)||p(z)), \lambda) - \E_{q_{\phi}} [\log p_{\theta}(x|z)]
\end{align*}

Seq2seq VAEs are prone to posterior collapse, so in practice, the rates obtained are very close to the target rates $\lambda$. 

As observed by \citet{alemi_fixing_2018}, different models or sets of hyperparameters for a given model can yield very similar values of ELBos despite reaching very different rates. 
Thus, for our purposes, the free bits modification is also useful to compare models with similar capacity.

\subsection{Variants}

Throughout the paper, we use variants of the original architecture and training procedure. In general, these variants use free bits objectives, but reach lower perplexities than what free bits alone allow.

\citet{li_surprisingly_2019}'s method is the following: pretrain an \textit{AE}, reinitialize the weights of the decoder, train the entire model again end-to-end with the VAE objective. The sentence representation $r$ is also the last hidden state of the LSTM encoder, so we call this method \textit{last-PreAE}. 

In the second variant, proposed by \citet{long_preventing_2019}, the representation of the document $r$ is the component-wise maximum over hidden states $h_i$, i.e., $r^j = \max_i h_i^j$. We call this model \textit{max}. In later experiments, we also consider a hybrid of the two techniques, \textit{max-PreAE}.

We chose these two baselines because they are relatively recent and outperform or perform on par with other recent methods such that cyclical learning rates \citep{fu_cyclical_2019} or aggressive training \citep{he_lagging_2019}. Moreover, the pooling encoder of \citet{long_preventing_2019} is particularly interesting: since pooling operators aggregate information over sets of vectors, they might prevent the copying of local information in the latent variable.

We make slight, beneficial modifications to these two methods. We remove KL annealing, which is not only redundant with the free bits technique but also increases the rate erratically \citep{pelsmaeker_effective_2019}. Moreover, for \citet{li_surprisingly_2019}'s method, we use $\delta$-VAE-style free bits instead of the original free bits to get rid of the unnecessary constraint that the free bits be balanced across components. For more details, see Appendix \ref{appendix:modif_to_li}. In summary, all of our experiments use $\delta$-VAE-style free bits without KL annealing.

Finally, \textit{AE} denotes the deterministic autoencoder trained only with the reconstruction loss.

\subsection{Datasets}

We train VAEs on four small versions of the AGNews, Amazon, Yahoo, and Yelp datasets created by \citet{zhang_character-level_2015}. Each document is written in English and consists of one or several sentences. Each document is labeled manually according to its main topic or the sentiment it expresses, and the labels are close to uniformly balanced over all the datasets. For faster training, we use smaller datasets. The characteristics of these datasets are detailed in Table \ref{table_datasets} in the Appendix.

\section{Encoders partially memorize the first words and sentence length}
\label{diagnosis}

The ELBo objective trades off the KL term against the reconstruction term. To minimize the objective, it is worth increasing the KL term only if the reconstruction term is decreased by the same amount or more. With free bits, the encoder is allowed to store a fixed amount of information for free. The objective becomes to minimize the reconstruction cost using the ``free storage'' as efficiently as possible. 

There are many solutions to this objective that are undesirable. For instance, we could program an encoder that would encode the words into the latent variable losslessly, until all the free bits are used. However, this model would not be more useful than a standard, left-to-right autoregressive models. Therefore, it is necessary to check that such useless, purely local features are not learned. 

In order to visualize \textit{what} information is stored in the latents, our method is to look at \textit{where} gains are seen in the reconstruction loss. Since the loss is a sum over documents and positions in these documents, these gains could be concentrated: i) on certain documents, for example, on large documents or documents containing rarer words; ii) at certain positions in the sentence, for example, in the beginning or in the middle of the sentence. We investigate the latter possibility.\footnote{PyTorch \citep{pytorch} implementation available at \url{https://github.com/tombosc/exps-s2svae}.}

\subsection{Visualizing the reconstruction loss}

Concretely, we compare the reconstruction loss of different models at different positions in the sentence. The baseline is a LSTM trained with a language model objective (\textit{LSTM-LM}). It has the same size as the decoders of the autoencoder models.\footnote{Only the input dimensions slightly change because in VAEs, the inputs of the decoder also include the latent vector.} Since the posterior collapse makes VAEs behave exactly like the \textit{LSTM-LM}, the reconstruction losses between the VAEs and the \textit{LSTM-LM} are directly comparable. Additionally, the deterministic \textit{AE} gives us the reconstruction error that is reachable with a latent space constrained only by its dimension $d$, but not by any target rate $\lambda$ (equivalent to an infinite target rate).

\begin{figure*}
\vspace{-0.15cm}
\begin{center}
  \includegraphics[scale=0.55]{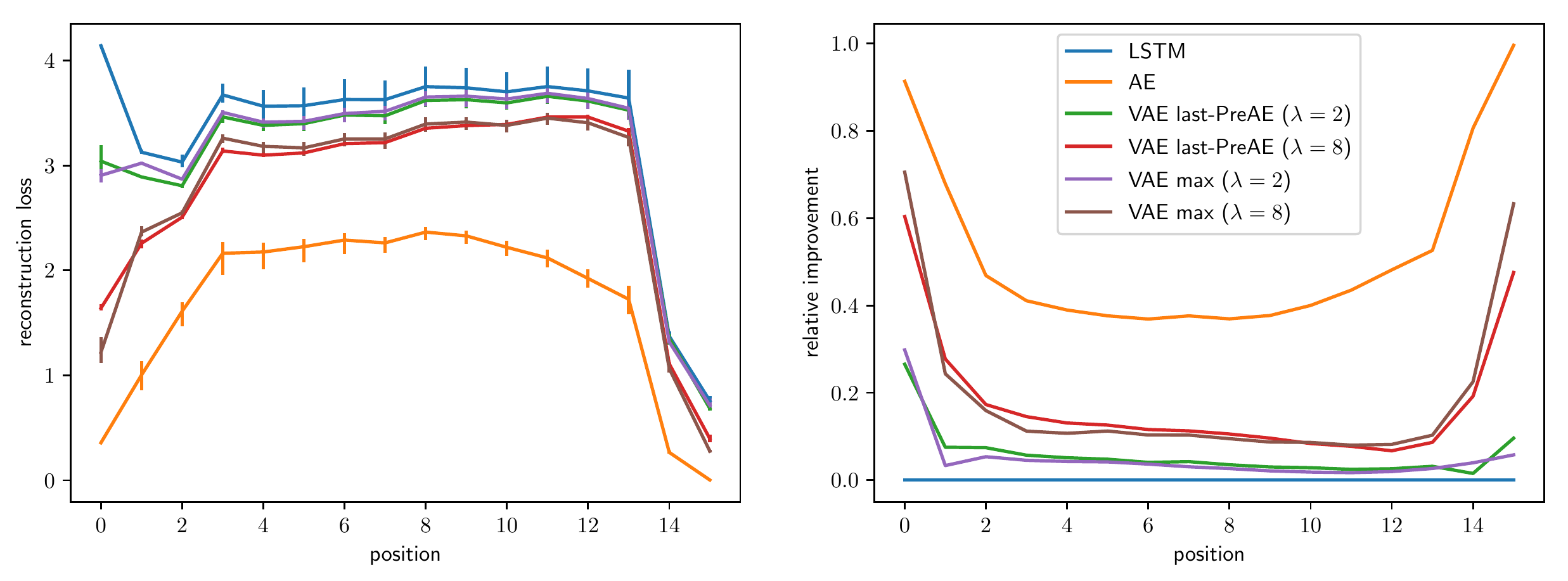}
\end{center}
\vspace{-0.5cm}
    \caption{\textit{Left}: Reconstruction loss on Yahoo dataset per each position in the sentence, averaged over sentences of 15 words (error bars: min, max on 3 runs); \textit{Right}: Relative improvement compared to baseline LSTM. Seq2seq autoencoders consistently store information about the first couple of words as well as the sentence length in priority.}
	\label{fig:loss_yahoo}
\end{figure*}

In Figure \ref{fig:loss_yahoo}, the left-hand side plot shows the reconstruction losses of different models and different target rates $\lambda$ on the Yahoo dataset. As expected, for all models, raising the target rate lowers the reconstruction cost.
Remarkably, these gains are very focused around the beginning and the end of documents. For a clearer picture of the gains at the end of the sentence, we plot the \textit{relative improvement} in reconstruction with respect to the baseline (right-hand side of Figure \ref{fig:loss_yahoo}) using:

$$\tilde{r}(i) = \frac{\max(r_{\mathrm{LSTM}}(i) - r(i), 0)}{r_{\mathrm{LSTM}}(i)}$$

where $r_{\mathrm{LSTM}}(i)$ is the loss of the LSTM. 

All the models reconstruct the first couple of words and the penultimate token better than the \textit{LSTM-LM}. On the three other datasets, there are similar peaks on relative improvements in the beginning and the end of sentences (Appendix \ref{app_other_plots}). 

It is not obvious that a lower reconstruction at a given position corresponds to information stored about the word in that position in the latent vector. Indeed, words are not independently modeled. However, we argue that it is roughly the case because the decoder is factorised from left-to-right and because correlations between words decrease with their distance in the sentence. The argument is detailed in the Appendix \ref{app_tracing_back}. 

How much do these gains on the reconstruction loss translate to decoding the first words and the document lengths more accurately? To find out, we compare regular VAEs to fixed-encoder, ideal VAEs that encode the true label perfectly and exclusively (in other words, VAEs whose latent variable is the ground-truth label). On sentence reconstruction, we found that regular VAEs decoded the first word 2 to 5 times more often than the baselines, indicating memorization of the first word. We also found similar but less dramatic results for sentence length (see Appendix \ref{app_decoding} for details).

This phenomenon was already noticed by \citet{kim_semi_2018}, using a different method (saliency measures, see Appendix \ref{sec:rel_eval} for details).

To sum up, compared to an unconditional \textit{LSTM-LM}, the seq2seq VAEs incur a much lower reconstruction loss on the first tokens and towards the end of the sentence (around 50\% less with $\lambda=8$). Moreover, if the latent variable of the VAEs did encode the label perfectly and exclusively, they would reconstruct the first words or recover sentence length with much lower accuracy than what is observed. Therefore, we conclude that seq2seq VAEs are biased towards memorizing the first few words and the sentence length. 

\subsection{The problem with memorization}

One could argue that this is a superficial problem, as we can always give the model more free bits and decrease the loss in intermediary positions. However, this is not so simple because increasing capacity leads to a worse model fit, as was noted by \citet{alemi_fixing_2018}. 
More specifically, on text data, \citet{prokhorov_importance_2019} noted that the coherence of samples decreases as the target rate increases. \citet{pelsmaeker_effective_2019} reported similar findings, and also, that more complex priors or posteriors do not help. Therefore, given current techniques, higher rates come at the cost of worse modeling of the data and therefore, we should strive for latent-variable models that store less information, but more global information.

Secondly, for controllable generation, conditioning on memorized information is useless. 
When the first words are encoded in the latent variable, the factorization of the VAE becomes the same as that of the usual autoregressive models, which are naturally able to continue a given beginning of the sentence (a ``prompt''). Similarly, document length is easily controlled by stopping the sampling after producing the desired number of words.\footnote{Or by explicitly conditioning on the sentence length. It can be useful for unsupervised summarization \citep{schumann_unsupervised_2018}, in flow-based approaches \citep{ziegler_latent_2019}, or more broadly for the decoder to plan sentence construction.} 
Finally, even for semi-supervised learning, a classifier that would only use the first few words and the sentence length would be suboptimal.

If these arguments are correct, it is doubtful that common seq2seq VAE architectures and training procedures in the low-capacity regime would learn useful representations. This is precisely the third problem: most of the KL values reported in the literature are low.\footnote{Most papers do not report the log base ($1$ bit is $\ln(2) \approx 0.693$ nats). Here are some reported rates of the best models: \citet{bowman_large_2015}: 2.0 (PTB) ; \citet{long_preventing_2019}: 3.7 (Yahoo), 3.1 (Yelp); \citet{li_surprisingly_2019}: 15.02 (Yahoo), 8.15 (PTB); \citet{he_lagging_2019}: 5.6 (Yahoo), 3.4 (Yelp); \citet{fu_cyclical_2019}: 1.955 (PTB), ...} Therefore, it is not clear whether the reported gains in performance (however measured) are significant, and if they are, what exactly cause these gains.

\section{Improving existing models}
\label{variants}

What architectures could avoid learning to memorize? We investigate simple variants and for a more thorough comparison with existing models, we refer to Appendix \ref{section_related_models}.

Our first variant uses a simple bag-of-words (\textit{BoW}) encoder in place of the LSTM encoder. The sentence representation is $r^j = \max_i e^j_i$, where the exponents denote components, and the indices denote positions in the sentence. We call it \textit{BoW-max-LSTM}. It is similar to the max-pooling model of \citet{long_preventing_2019} except that the maximum is taken over embeddings rather than LSTM hidden states. As \citet{long_preventing_2019} reported, the max-pooling operator is better than the average operator, both when the encoder is a LSTM and \textit{BoW} (possibly because the maximum introduces a non-linearity). Therefore, we use the maximum operator. A priori, we think that since word order is not provided to the encoder, the encoder should be unable to memorize the first words.

For our second variant we use a unigram decoder (\textit{Uni}) in place of an LSTM decoder. It produces a single output probability distribution  for all positions in the sentence $i$, conditioned only on the latent variable $z$. This distribution is obtained by applying a one-hidden layer MLP followed by softmax to the latent vector: $p_{\theta}(x_i|z) = \softmax(W_2 \mathrm{ReLU}(W_1z) + b)$, where $\mathrm{ReLU}(x) = \max(x,0)$ \citep{nair2010rectified}. We hope that the encoder will learn representations that do not focus on the first words, because the decoder should not need this particular information. We can use any encoder in combination of this decoder and if we use a \textit{BoW} encoder, we obtain the NVDM model of \citet{miao_neural_2016}.

Both the \textit{BoW} encoders and \textit{Uni} decoders variants might benefit from the \textit{PreAE} pretraining technique, but we leave this for future work. 

Lastly, the pretrained LM (\textit{PreLM}) variant is obtained in two training steps. First, we pretrain a \textit{LSTM-LM}. Then, it is used as an encoder with fixed weights. We use average pooling over the hidden states to get a sentence representation, i.e., $r = \frac{1}{L} \sum_{i=1}^L h_i$, and learn the transformations $L_1$ and $L_2$ that compute the variational parameters. Initially, we tried to use max-pooling but the training was extremely unstable. The LM objective requires the hidden state to capture both close correlations between words but also more global information to predict long-distance correlations. The hope is that this global information can be retrieved via pooling and encoded in the variational parameters. The \textit{PreLM} variant is nothing more than the use of a pretrained LM as a feature extractor \citep{peters_deep_2018}. While \citet{yang_improved_2017} and \citet{kim_semi_2018} both consider the use of pretrained LMs as encoders, the weights are not frozen such that it is hard to disentangle the impact of pretraining from subsequent training. In contrast, we freeze the weights so that the effect of pretraining can not be overridden.  
To isolate the effect of this training procedure independently of the architecture, we keep the same LSTM instead of using more powerful architectures such as Transformers.

\section{Semi-supervised learning evaluation}
\label{section_ssl}

We turn to the semi-supervised learning (\textit{SSL}) setting to compare the learned representations of our variants. For the purpose of controllable text generation, we assume that the global information that is desirable to capture is the topic or sentiment. There are two training phases: first, an unsupervised pretraining phase where VAEs are trained; second, a supervised learning phase where classifiers are trained to predict ground-truth labels given the latent vectors encoded with the encoders of the VAEs. This is essentially the same setup as \textit{M1} from  \citet{kingma_semi-supervised_2014}.\footnote{We could integrate the labels into the generative model as a random variable that is either observed or missing to obtain better results \citep{kingma_semi-supervised_2014}. Still, our goal is to study the inductive bias of the seq2seq VAE \textit{as an unsupervised learning method}, so we do not train the encoder using the labels.} The small and large data-regimes give us complementary information: with many labels and complex classifiers, we quantify \textit{how much} of the information pertaining to the labels is encoded; with few labels and simple classifiers, \textit{how accessible} the information is. 

For each dataset, we subsample $g=5$ balanced labeled datasets for each different data-regimes, containing 5, 50, 500, and 5000 examples per class. These labeled datasets are used for training and validating during the supervised learning phase.\footnote{It is especially important to use several subsamples in the low data-regimes where subsamples containing unrepresentative texts or noisy labels are not unlikely.} Each model is trained with $s=3$ seeds. The performance of the classifiers are measured by the macro F1-score on the entire test sets.

To select hyperparameters on each subsample, we use repeated stratified K-fold cross-validation \citep{moss_using_2018} as detailed in the Appendix \ref{app_cv}. We obtain the test set F1-scores $F_{ij}$, where $i$ is the subsample seed and $j$ is the parameter initialisation seed, and report $\bar{F_{\cdot \cdot}}$, the average F1-score over $i$ and $j$. We note $\bar{F_{ \cdot j}}$ the empirical average F1-score for a given parameter initialisation $j$ and decompose the variance into:

\begin{itemize}
    \item $\sigma_{\mathrm{init}} = (\frac{1}{s-1} \sum_{j=1}^s g  (\bar{F_{\cdot j}} - \bar{F_{\cdot \cdot}})^2)^{\frac{1}{2}}$, which quantifies the variability due to the initialisation of the model,
    \item $\sigma = (\frac{1}{g} \sum_{i=1}^g \frac{1}{s - 1} \sum_{j=1}^s (F_{ij} - \bar{F_{\cdot j}})^2)^{\frac{1}{2}}$, which quantifies the remaining variability.
\end{itemize}

In the context of ANOVA with a linear model and a single factor, these quantities are the square roots of $MS_T$ and $MS_E$ (see Appendix \ref{appendix:stats_ssl}).

Finally, we also add a data-regime where the entire labeled training set is used in the supervised learning phase. In this setting, we use more expressive one-hidden-layer MLP classifiers, with early stopping on the validation set. Thus, we can check that our conclusions in the large data-regime do not depend on the model selection procedure and the choice of the classifier.

For each class of model, we perform a grid search over target rates and latent vector sizes. We search for target rates $\lambda$ in $\{2,8\}$: large enough to capture label information but small enough to avoid underfitting, as explained above. The size of latent vectors $d$ are chosen in $\{4,16\}$. They should be small enough for extremely low-data regimes. For instance, on Yelp, the smallest data regime (5 per class) uses only 8 examples to train the classifier and 2 to do cross-validation. A thorough explanation is presented in Appendix \ref{appendix:training}, along with the values of hyperparameters held constant.

What representation should be used as inputs to the classifiers? \citet{kingma_improved_2016} use samples from the approximate posterior $q_{\phi}(z|x) = \mathcal{N}(z|\mu, \mathrm{diag}(\sigma^2))$, but in the NLP literature, most evaluations focus on $\mu$ without mention or justification. To evaluate the VAE as a generative model, we claim that only noisy samples $z$ should be used. In fact, using a model with a rate close to 0 on Yelp, we can recover the label with a high F1-score of $81.5\%$ by using $\mu$, whereas, as expected, noisy samples $z$ do not do better than random ($50\%$). The information contained in $\mu$ is misleading because it is not transmitted to the decoder and not used directly during generation. Therefore, we use samples $z$ (cf. Appendix \ref{section_mean_vs_samples} for details).

\begin{table}[t]

\begin{center}
\begin{small}
\begin{adjustbox}{max width=0.50\textwidth}
	\begin{tabular}{l|l@{~~}l@{~~}l@{~~}l|lHHHHHl}
\toprule                  
 & & & & &                5 &             10 &             50 &            500 &           5000 & All & All   \\
 & Enc. & $r$ & Dec. & Pre. &                \multicolumn{7}{c}{F1~$^{\sigma}_{\sigma_{\mathrm{init}}}$} \\
 \midrule
 \multirow{7}{*}{\rotatebox[origin=c]{90}{AGNews}} & LSTM &  last &  LSTM &  AE &   $65.8$~$^{3.3}_{3.3}$ &   $73.9$~$^{4.2}_{6.6}$ &   $81.0$~$^{0.7}_{1.1}$ &   $82.8$~$^{0.3}_{0.6}$ &   $83.1$~$^{0.1}_{0.7}$ &   $82.7$~$^{-}_{0.4}$ &   $83.4$~$^{-}_{0.3}$ \\
 & LSTM &   max &  LSTM &  AE &  $55.7$~$^{4.5}_{18.7}$ &  $64.8$~$^{6.0}_{15.5}$ &   $75.1$~$^{1.3}_{2.6}$ &   $81.9$~$^{0.3}_{0.0}$ &   $82.5$~$^{0.1}_{0.4}$ &   $82.2$~$^{-}_{0.4}$ &   $83.3$~$^{-}_{0.4}$ \\
 & BoW &   max &  LSTM &      - &   $72.7$~$^{2.0}_{5.9}$ &    $77.1$~$^{2.2}_{3.2}$ &   $81.2$~$^{0.6}_{0.8}$ &   $82.2$~$^{0.2}_{0.8}$ &   $82.3$~$^{0.1}_{1.0}$ &   $82.0$~$^{-}_{0.3}$ &   $83.1$~$^{-}_{0.3}$ \\
 &LSTM &   max &   Uni &      - &   $71.6$~$^{5.5}_{0.1}$ &   $75.0$~$^{4.3}_{1.6}$ &   $80.4$~$^{0.8}_{0.7}$ &   $81.8$~$^{0.5}_{0.5}$ &   $82.4$~$^{0.1}_{0.4}$ &   $82.2$~$^{-}_{0.4}$ &   $83.9$~$^{-}_{0.3}$ \\
 &LSTM &  last &   Uni &      - &  $54.8$~$^{5.2}_{57.1}$ &  $58.2$~$^{3.3}_{65.6}$ &  $61.7$~$^{0.8}_{71.4}$ &  $62.9$~$^{0.4}_{71.0}$ &  $63.0$~$^{0.3}_{71.1}$ &  $61.5$~$^{-}_{34.5}$ &  $59.3$~$^{-}_{40.9}$ \\
 & BoW &   max &   Uni &      - &   $71.8$~$^{4.5}_{1.8}$ &   $75.6$~$^{2.6}_{1.6}$ &   $81.4$~$^{0.5}_{0.6}$ &   $82.5$~$^{0.1}_{0.5}$ &   $82.5$~$^{0.1}_{0.6}$ &   $82.3$~$^{-}_{0.3}$ &   $83.1$~$^{-}_{0.5}$ \\
 &LSTM &   avg &  LSTM &      LM &   $70.8$~$^{4.8}_{4.3}$ &   $76.3$~$^{3.0}_{2.7}$ &   $81.2$~$^{0.9}_{1.2}$ &   $82.6$~$^{0.2}_{1.3}$ &   $82.8$~$^{0.1}_{0.9}$ &   $82.5$~$^{-}_{0.4}$ &   $83.5$~$^{-}_{0.1}$ \\

\midrule
 \multirow{7}{*}{\rotatebox[origin=c]{90}{Amazon}} & LSTM &  last &  LSTM &  AE &  $20.0$~$^{2.2}_{0.9}$ &  $21.2$~$^{1.4}_{1.5}$ &  $24.7$~$^{0.7}_{2.8}$ &  $27.2$~$^{0.4}_{3.1}$ &  $27.7$~$^{0.3}_{3.8}$ &  $28.1$~$^{-}_{1.1}$ &  $28.1$~$^{-}_{1.0}$ \\
 & LSTM &   max &  LSTM &  AE &  $22.3$~$^{2.6}_{0.7}$ &  $24.9$~$^{3.8}_{1.5}$ &  $30.5$~$^{0.9}_{3.0}$ &  $33.4$~$^{0.4}_{4.1}$ &  $34.1$~$^{0.3}_{4.8}$ &  $34.0$~$^{-}_{1.6}$ &  $34.0$~$^{-}_{1.6}$ \\
 &  BoW &   max &  LSTM &      - &  $21.0$~$^{2.6}_{1.1}$ &  $27.6$~$^{3.0}_{1.2}$ &  $34.6$~$^{0.7}_{1.1}$ &  $38.3$~$^{0.4}_{1.0}$ &  $39.0$~$^{0.1}_{0.6}$ &  $39.2$~$^{-}_{0.6}$ &  $38.9$~$^{-}_{0.7}$ \\
 & LSTM &   max &   Uni &      - &  $21.8$~$^{3.1}_{1.6}$ &  $25.9$~$^{3.1}_{1.3}$ &  $32.8$~$^{0.8}_{1.7}$ &  $36.9$~$^{0.4}_{0.9}$ &  $38.0$~$^{0.2}_{0.6}$ &  $36.8$~$^{-}_{0.7}$ &  $38.2$~$^{-}_{0.5}$ \\
 & LSTM &  last &   Uni &      - &  $24.0$~$^{3.0}_{1.0}$ &  $28.2$~$^{1.9}_{3.3}$ &  $31.2$~$^{0.6}_{1.4}$ &  $35.1$~$^{0.4}_{2.2}$ &  $36.1$~$^{0.2}_{2.4}$ &  $36.3$~$^{-}_{1.1}$ &  $36.8$~$^{-}_{0.9}$ \\
 & BoW &   max &   Uni &      - &  $25.4$~$^{3.2}_{0.2}$ &  $29.0$~$^{2.4}_{2.8}$ &  $32.8$~$^{1.0}_{1.3}$ &  $36.1$~$^{0.4}_{0.7}$ &  $36.9$~$^{0.2}_{0.8}$ &  $37.4$~$^{-}_{0.2}$ &  $37.9$~$^{-}_{0.2}$ \\
 & LSTM &   avg &  LSTM &      LM &  $21.8$~$^{3.8}_{0.6}$ &  $28.9$~$^{1.8}_{1.0}$ &  $35.3$~$^{0.8}_{0.4}$ &  $40.2$~$^{0.4}_{0.4}$ &  $41.1$~$^{0.2}_{0.4}$ &  $41.1$~$^{-}_{0.8}$ &  $40.0$~$^{-}_{0.4}$ \\

\midrule
 \multirow{7}{*}{\rotatebox[origin=c]{90}{Yahoo}} &LSTM &  last &  LSTM &  AE &  $20.7$~$^{0.7}_{0.5}$ &  $23.1$~$^{1.6}_{1.6}$ &  $32.2$~$^{0.8}_{0.6}$ &  $36.1$~$^{0.2}_{0.1}$ &  $36.7$~$^{0.1}_{0.5}$ &  $36.5$~$^{-}_{0.1}$ &  $37.2$~$^{-}_{0.7}$ \\
 &LSTM &   max &  LSTM &  AE &  $20.8$~$^{1.3}_{2.3}$ &  $23.2$~$^{1.6}_{1.2}$ &  $31.3$~$^{0.7}_{1.4}$ &  $35.6$~$^{0.3}_{1.2}$ &  $36.3$~$^{0.1}_{1.1}$ &  $36.0$~$^{-}_{0.7}$ &  $36.6$~$^{-}_{0.7}$ \\
 & BoW &   max &  LSTM &      - &  $23.4$~$^{2.1}_{2.9}$ &  $26.2$~$^{2.1}_{1.2}$ &  $36.7$~$^{1.1}_{0.5}$ &  $41.1$~$^{0.2}_{0.8}$ &  $41.6$~$^{0.1}_{0.9}$ &  $41.6$~$^{-}_{0.3}$ &  $42.6$~$^{-}_{0.2}$ \\
 &LSTM &   max &   Uni &      - &  $24.9$~$^{1.3}_{2.2}$ &  $25.9$~$^{2.5}_{0.7}$ &  $33.2$~$^{0.7}_{3.6}$ &  $37.3$~$^{0.1}_{3.1}$ &  $37.9$~$^{0.1}_{3.1}$ &  $37.6$~$^{-}_{1.4}$ &  $38.9$~$^{-}_{1.7}$ \\
 &LSTM &  last &   Uni &      - &  $24.5$~$^{3.8}_{1.7}$ &  $27.6$~$^{2.2}_{1.5}$ &  $30.8$~$^{1.7}_{0.6}$ &  $34.4$~$^{0.3}_{5.0}$ &  $35.1$~$^{0.1}_{4.7}$ &  $35.1$~$^{-}_{1.9}$ &  $37.1$~$^{-}_{2.3}$ \\
 & BoW &   max &   Uni &      - &  $24.1$~$^{2.9}_{2.7}$ &  $26.0$~$^{5.0}_{1.8}$ &  $35.0$~$^{0.9}_{1.2}$ &  $39.1$~$^{0.1}_{1.8}$ &  $39.5$~$^{0.1}_{1.7}$ &  $39.2$~$^{-}_{0.4}$ &  $40.1$~$^{-}_{0.7}$ \\
 &LSTM &   avg &  LSTM &      LM &  $21.9$~$^{2.3}_{1.3}$ &  $26.1$~$^{2.0}_{1.1}$ &  $36.1$~$^{0.8}_{0.7}$ &  $39.9$~$^{0.2}_{0.6}$ &  $40.4$~$^{0.1}_{0.4}$ &  $40.9$~$^{-}_{0.4}$ &  $41.7$~$^{-}_{0.3}$ \\

\midrule
 \multirow{7}{*}{\rotatebox[origin=c]{90}{Yelp}} & LSTM &  last &  LSTM &  AE &    $59.3$~$^{5.4}_{2.9}$ &   $65.6$~$^{4.9}_{3.2}$ &  $80.0$~$^{1.3}_{3.0}$ &  $82.7$~$^{1.0}_{0.9}$ &  $83.3$~$^{0.1}_{2.3}$ &  $83.5$~$^{-}_{0.8}$ &  $67.9$~$^{-}_{0.1}$ \\
 & LSTM &   max &  LSTM &  AE &   $59.9$~$^{10.4}_{7.9}$ &   $62.9$~$^{4.4}_{2.8}$ &  $78.7$~$^{2.4}_{1.5}$ &  $82.9$~$^{0.3}_{2.7}$ &  $83.3$~$^{0.1}_{2.7}$ &  $83.6$~$^{-}_{0.7}$ &  $84.1$~$^{-}_{0.7}$ \\
 & BoW &   max &  LSTM &      - &  $67.1$~$^{10.1}_{15.7}$ &  $70.8$~$^{6.8}_{13.0}$ &  $79.3$~$^{2.8}_{4.5}$ &  $83.4$~$^{0.3}_{0.9}$ &  $83.9$~$^{0.1}_{0.9}$ &  $84.2$~$^{-}_{0.8}$ &  $85.0$~$^{-}_{0.2}$ \\
 &LSTM &   max &   Uni &      - &    $62.3$~$^{4.6}_{3.8}$ &   $68.5$~$^{4.7}_{4.3}$ &  $76.7$~$^{1.7}_{3.6}$ &  $80.4$~$^{0.2}_{3.2}$ &  $80.9$~$^{0.1}_{3.1}$ &  $81.1$~$^{-}_{0.9}$ &  $83.1$~$^{-}_{0.5}$ \\
 & LSTM &  last &   Uni &      - &    $65.0$~$^{8.0}_{4.4}$ &   $68.8$~$^{4.0}_{3.9}$ &  $74.1$~$^{2.0}_{1.4}$ &  $78.5$~$^{0.3}_{3.0}$ &  $79.1$~$^{0.1}_{3.1}$ &  $79.3$~$^{-}_{1.1}$ &  $81.6$~$^{-}_{0.5}$ \\
 & BoW &   max &   Uni &      - &    $59.9$~$^{7.2}_{3.7}$ &   $68.8$~$^{5.0}_{2.8}$ &  $77.3$~$^{1.2}_{0.9}$ &  $81.1$~$^{0.3}_{0.5}$ &  $81.5$~$^{0.1}_{0.5}$ &  $81.4$~$^{-}_{0.1}$ &  $83.3$~$^{-}_{0.4}$ \\
 & LSTM &   avg &  LSTM &      LM &    $63.6$~$^{7.4}_{5.4}$ &   $71.0$~$^{6.8}_{2.7}$ &  $81.0$~$^{1.6}_{2.3}$ &  $83.2$~$^{0.7}_{0.8}$ &  $83.8$~$^{0.1}_{0.8}$ &  $83.6$~$^{-}_{0.7}$ &  $84.4$~$^{-}_{0.5}$ \\
\bottomrule
\end{tabular}
\end{adjustbox}
\end{small}
\end{center}
	\caption{Using \textit{BoW} encoders, \textit{Uni} decoders or \textit{PreLM} pretraining, the learned representations are more predictive of the labels (sentiment or topic).}
\label{table_pooling_small}
\end{table}

\subsection{Results}

Table \ref{table_pooling_small} contains the results of the SSL experiments in the smallest and largest data-regimes. The results for intermediary data-regimes as well as for baseline models without pretraining, which underperform, are presented in the Appendix, Table \ref{table_pooling}. The proposed variants are either on par or improve significantly over the baselines. In the large data-regime, \textit{BoW-max-LSTM} and \textit{LSTM-avg-LSTM-PreLM} perform best on average while \textit{LSTM-last-Uni} performs the worst and suffers from unstable training on AGNews. In the small data-regime, the picture is less clear because there is more variance.

On AGNews and Yelp, in the large data-regime, our variants do not seem to improve over the baselines. However, on Amazon and Yahoo, in the large data-regime, the variants seem to improve by $5$ in F1-score. Why do the gains vary so widely depending on the datasets? We posit that, on some datasets, the first words are enough to predict the labels correctly. We train bag-of-words classifiers
\footnote{fastText classifiers \citep{joulin_bag_2017} with embedding dimension of 200 and the default parameters.}
using either i) only the first three words or ii) all the words as features on the entire datasets. If the \textit{three-words} classifiers are as good as the \textit{all-words} classifiers, we expect that the original VAE variants will perform well: in that case, encoding information about the first words is not harmful, it could be a rather useful inductive bias. Conversely, if the first three words are not predictive of the label, the original VAEs will perform badly. 

As reported in the Appendix, Table \ref{table_bow_all_vs_three}, on AGNews and Yelp, classifiers trained on the first three words have a performance somewhat close to the classifier trained on all the words, reaching 80.8\% and 85.4\% of its scores respectively. For instance, on AGNews, the first words are often nouns that directly gives the topic of the news item: country names for the politics category, firm names for the technology category, athlete or team names for the sports category, etc. On the two other datasets, the performance decays a lot if we only use the first three words: \textit{three-words} F1-scores make up for 60.7\% and 30.3\% of \textit{all-words} F1-scores on Amazon and Yahoo. This explains why the original VAE can perform on par or slightly better than our variants on certain datasets for which the first words are very predictive of the labels. This also proves that using several datasets is necessary to draw robust conclusions.

Despite similar asymptotic performance on AGNews and Yelp, our variants clearly improve over the baselines in the small data-regime, which suggests that the encoded information is quantitatively different. This is confirmed in the next section.

It might be surprising that \textit{LSTM-max-LSTM} models are inferior to \textit{BoW-max-LSTM} models. In Appendix \ref{app_workaround}, we show that with recurrent encoders, some components of the hidden states are consistently maximized at certain early positions in the sentence. This explains why the power of LSTMs can be undesirable, and why the simpler \textit{BoW} encoders perform better.

\section{Text generation evaluation}
\label{section_text_gen}

\begin{table}
\begin{center}
\begin{adjustbox}{max width=0.47\textwidth}
	\begin{tabular}{Hl@{~~}l@{~~}l@{~~}|l@{~~}l@{~~}Hl@{~}l@{~}}
\toprule

& Enc. & $r$ & Pre. &  Agree. & 1st (\%) &    Mid (\%) & Len (\%) &        $\approx$PPL \\
\midrule
\multirow{5}{*}{\rotatebox[origin=c]{90}{AGNews}} & LSTM & last & AE &  $80.2${\footnotesize $\pm1.0$} &         $29.6${\footnotesize $\pm1.1$} &  $2.7${\footnotesize $\pm1.0$} &      $3.6${\footnotesize $\pm0.1$} &  \boldmath $34.8${\footnotesize $\pm0.4$} \\
& LSTM & max  & AE &  $79.5${\footnotesize $\pm0.9$} &         $31.7${\footnotesize $\pm1.1$} &  $2.7${\footnotesize $\pm1.1$} &      $3.7${\footnotesize $\pm0.5$} &  \boldmath $34.7${\footnotesize $\pm0.4$} \\
& BoW & max  & - &  $78.0${\footnotesize $\pm1.3$} &         $18.9${\footnotesize $\pm1.2$} &  $2.2${\footnotesize $\pm0.6$} &     \boldmath $2.7${\footnotesize $\pm0.3$} &  $36.1${\footnotesize $\pm0.6$} \\
& BoW & max  & Uni &   $81.3${\footnotesize $\pm0.1$} &        \boldmath $13.9${\footnotesize $\pm0.3$} &  \boldmath $2.9${\footnotesize $\pm0.8$} &      $3.1${\footnotesize $\pm0.1$} &  $36.3${\footnotesize $\pm0.7$} \\
& LSTM & max & Uni &  \boldmath $82.0${\footnotesize $\pm0.4$} &         \boldmath $13.9${\footnotesize $\pm0.2$} &  $2.9${\footnotesize $\pm0.9$} &      $3.3${\footnotesize $\pm0.4$} &  $36.0${\footnotesize $\pm0.4$} \\
& LSTM & avg & LM &  $79.2${\footnotesize $\pm0.4$} &         $22.2${\footnotesize $\pm0.8$} &  $2.7${\footnotesize $\pm0.8$} &      $3.2${\footnotesize $\pm0.2$} &  \boldmath $35.0${\footnotesize $\pm0.3$} \\
\midrule
\multirow{5}{*}{\rotatebox[origin=c]{90}{Amazon}} & LSTM & last & AE &  $24.5${\footnotesize $\pm0.4$} &         $42.4${\footnotesize $\pm2.3$} &  $2.9${\footnotesize $\pm0.8$} &     $13.0${\footnotesize $\pm1.6$} &  \boldmath $44.5${\footnotesize $\pm0.2$} \\
& LSTM & max &  AE &  $30.8${\footnotesize $\pm1.1$} &         $41.7${\footnotesize $\pm0.8$} &  $2.8${\footnotesize $\pm0.6$} &     $11.5${\footnotesize $\pm1.0$} &  \boldmath $44.4${\footnotesize $\pm0.3$} \\
& BoW & max & - &  $34.2${\footnotesize $\pm0.5$} &         $33.3${\footnotesize $\pm0.7$} &  $2.6${\footnotesize $\pm0.4$} &     \boldmath $9.9${\footnotesize $\pm0.7$} &  $45.3${\footnotesize $\pm0.5$} \\
& BoW & max &  Uni &  $33.3${\footnotesize $\pm0.4$} &        \boldmath $21.5${\footnotesize $\pm0.3$} &  $2.7${\footnotesize $\pm0.5$} &     $11.8${\footnotesize $\pm0.5$} &  $45.3${\footnotesize $\pm0.4$} \\
& LSTM & max & Uni &  $34.1${\footnotesize $\pm0.5$} &          $22.1${\footnotesize $\pm0.1$} &  $2.7${\footnotesize $\pm0.5$} &     $11.7${\footnotesize $\pm0.6$} &  $45.4${\footnotesize $\pm0.6$} \\

& LSTM & avg &  LM &  \boldmath $35.8${\footnotesize $\pm0.4$} &         $38.3${\footnotesize $\pm0.9$} &  \boldmath $3.0${\footnotesize $\pm0.7$} &     $11.5${\footnotesize $\pm1.0$} &  \boldmath $44.2${\footnotesize $\pm0.4$} \\
\midrule
\multirow{5}{*}{\rotatebox[origin=c]{90}{Yahoo}} & LSTM & last &  AE &  $23.8${\footnotesize $\pm0.2$} &         $56.6${\footnotesize $\pm1.0$} &  $6.3${\footnotesize $\pm0.9$} &     $17.1${\footnotesize $\pm1.1$} &   \boldmath $48.8${\footnotesize $\pm0.2$} \\
& LSTM & max &  AE &  $22.9${\footnotesize $\pm0.8$} &         $58.7${\footnotesize $\pm1.7$} &  $6.5${\footnotesize $\pm0.9$} &     $18.4${\footnotesize $\pm0.8$} &  \boldmath $48.6${\footnotesize $\pm0.1$} \\
& BoW & max &  - &  \boldmath $26.9${\footnotesize $\pm0.5$} &         $49.3${\footnotesize $\pm1.2$} &  $5.6${\footnotesize $\pm1.2$} &     $11.8${\footnotesize $\pm0.3$} &   $49.7${\footnotesize $\pm0.4$} \\
& BoW & max &  Uni &  \boldmath $26.8${\footnotesize $\pm0.6$} &         \boldmath $37.6${\footnotesize $\pm0.9$} &  $5.8${\footnotesize $\pm1.2$} &    \boldmath $10.6${\footnotesize $\pm0.4$} &   $49.8${\footnotesize $\pm0.1$} \\
& LSTM & max & Uni &  \boldmath $27.1${\footnotesize $\pm1.0$} &         \boldmath $37.7${\footnotesize $\pm1.6$} &  $6.1${\footnotesize $\pm1.3$} &     \boldmath $11.0${\footnotesize $\pm0.3$} &  $50.0${\footnotesize $\pm0.4$} \\

& LSTM & avg &  LM &  \boldmath $26.7${\footnotesize $\pm0.2$} &         $51.9${\footnotesize $\pm0.5$} &  \boldmath $6.7${\footnotesize $\pm0.9$} &     $16.7${\footnotesize $\pm1.8$} & \boldmath $48.5${\footnotesize $\pm0.1$} \\
\midrule
\multirow{5}{*}{\rotatebox[origin=c]{90}{Yelp}} & LSTM & last &  AE &  \boldmath $81.7${\footnotesize $\pm1.3$} &         $53.0${\footnotesize $\pm0.5$} &  \boldmath $8.5${\footnotesize $\pm2.0$} &     $33.7${\footnotesize $\pm1.7$} &  \boldmath $31.7${\footnotesize $\pm0.3$} \\
& LSTM & max &  AE &  $81.3${\footnotesize $\pm0.7$} &         $52.4${\footnotesize $\pm0.5$} &  $8.3${\footnotesize $\pm2.2$} &     $29.5${\footnotesize $\pm2.5$} &  \boldmath $31.8${\footnotesize $\pm0.1$} \\
& BoW & max &  - &  \boldmath $82.2${\footnotesize $\pm0.5$} &         $36.4${\footnotesize $\pm0.3$} &  $7.6${\footnotesize $\pm2.0$} &     $22.4${\footnotesize $\pm0.5$} &  $32.3${\footnotesize $\pm0.4$} \\
& BoW & max &  Uni &  $80.4${\footnotesize $\pm0.4$} &         \boldmath $30.6${\footnotesize $\pm0.5$} &  $6.3${\footnotesize $\pm1.7$} &    \boldmath $15.4${\footnotesize $\pm0.4$} &  $32.8${\footnotesize $\pm0.1$} \\
& LSTM & max & Uni &  $80.9${\footnotesize $\pm0.4$} &         $32.0${\footnotesize $\pm0.4$} &  $6.3${\footnotesize $\pm1.7$} &     $17.2${\footnotesize $\pm0.7$} &  $33.1${\footnotesize $\pm0.3$} \\

& LSTM & avg &  LM &  \boldmath $82.3${\footnotesize $\pm0.7$} &         $47.7${\footnotesize $\pm0.4$} &  $7.7${\footnotesize $\pm2.6$} &     $24.1${\footnotesize $\pm0.4$} &  \boldmath $31.9${\footnotesize $\pm0.2$} \\
\bottomrule
                                                   
    \end{tabular}
\end{adjustbox}
\end{center}
\caption{Our variants reconstruct inputs with higher agreement, less memorization of the 1st words and lengths and a negligible loss in likelihood. Best score and scores within one standard deviation are bolded.}
\label{table_decoding}
\end{table}

How do these different variants perform during generation? We expect that the SSL classification performances would correlate with the abilities of the decoders to reconstruct documents that exhibit a similar global aspect than the encoded documents. 

To measure the agreement in label between the source document and its reconstruction, we adapt the evaluation procedure used by \citet{ficler_controlling_2017} so that no human annotators or heuristics are required (see Appendix \ref{sec:rel_eval}). First, a classifier is trained to predict the label on the source dataset. Then, for each model, we encode the documents, reconstruct them, and classify these reconstructions using the classifier. The \textit{agreement} is the F1-scores between the original labels and the labels given by the classifiers on the generated samples. 

To quantify memorization, we measure the reconstruction accuracy of the first word and the ratio of identical sentence length between sources and reconstructions. Finally, to verify that our bag-of-words assumptions do not hurt the overall fit to the data, we estimate the negative log-likelihood via the importance-weighted lower bound \citep{burda_importance_2015} (500 samples) to compute an approximate perplexity per word (\textit{$\approx$PPL}).

We use two decoding schemes: beam search with a beam of size 5 and greedy decoding. We fix $\lambda=8$, $d=16$ on all models, with three seeds. For the \textit{Uni} decoder, we drop \textit{LSTM-last-Uni} which underperformed by a large margin in the SSL setting, and for the other \textit{Uni} models, we freeze the encoder, $L_1$ and $L_2$ and train a new recurrent decoder using the reconstruction loss only. Essentially, the \textit{Uni} decoder is an \textit{auxiliary decoder}, as described by \citet{de_fauw_hierarchical_2019} (see Appendix \ref{section_related_models} for details) and we denote this technique by \textit{PreUni}.

Table \ref{table_decoding} show the results for beam search decoding.\footnote{Similar results were obtained using greedy decoding, albeit sometimes consistently shifted.} There is a close correspondence between agreement and performance on the SSL tasks in the large data-regime. Our variants have a higher agreement than the baselines, especially on Amazon and Yahoo datasets for which the memorization of the first words is especially harmful.

The baselines reconstruct the first words with very high accuracy (more than 50\% of the time on Yahoo and Yelp) while our variants mitigate this memorization. For instance, \textit{PreUni} models recover the first word around 2 or 1.5 times less often. 

Let us focus on AGNews and Yelp, where the first words are very predictive of the labels. Both baselines and variants have roughly similarly high agreement. However, our variants produce more diverse beginnings, while still managing to reproduce the topic or sentiment of the original document. On the other hand, the reconstructions of the baselines exhibit the same labels as the sources mostly as a side-effect of starting with the same words. This also explains that in the SSL setting, despite similar performances asymptotically, our variants were much more efficient using five examples per class. Memorization of the first words does not abstract away from the particular words and therefore, the amount of data required to learn a good classifier will be high, compared to a model which truly \textit{infer} unobserved characteristics of documents.

Both \textit{BoW} encoders and \textit{Uni} decoders lower memorization, so bag-of-words assumptions are efficient for dealing with the memorization problem. Still, \textit{BoW-Max} and \textit{LSTM-Max} with \textit{PreUni} pretraining yield very close performance despite having a different encoder, showing that the decoder has a far greater influence than the encoder. This is consistent with \citet{mccoy_rnns_2019}'s findings (see Section \ref{sec:rel_eval} in Appendix for details).

Finally, there seems to be a trade-off between the global character of the latent information and the fit of the model to the data. \textit{BoW} and \textit{Uni} variants have perplexity roughly one unit above the baselines, a significant but small difference.

In Appendix \ref{app_qualitative}, we perform a qualitative analysis of reconstruction samples to illustrate these conclusions. It also sheds light on the inherent difficulty of the Yahoo dataset.

To recapitulate, the bag-of-words assumptions decrease the memorization of the first word and of the sentence length in the latent variable while increasing the agreement between the labels of the source and of the reconstruction. This is achieved at the cost of a small increase in perplexity.

\section{Conclusion}

Eliminating posterior collapse is necessary to get useful VAE models, but not sufficient. 
Although recent incarnations of the seq2seq VAE fix the posterior collapse, they partially memorize the first few words and the document lengths. Depending on the data, these local features are sometimes not very correlated with global aspects like topic or sentiment. Therefore, they are of limited use for controllable and diverse text generation.

To learn to infer more global features, we explored alternative architectures based on bag-of-word assumptions on the encoder or decoder side, as well as a pretraining procedure. These variants are all effective, in particular, the unigram decoder used as an auxiliary decoder \citep{de_fauw_hierarchical_2019}.
The latent variable is more predictive of global features and memorisation of the first words and sentence length is decreased. Thus, these models are more suitable for diverse and controllable generation.

Methodologically, we introduced a simple way to examine the content of latent variables by looking at the reconstruction loss per position. We also presented a reliable way to perform semi-supervised learning experiments to analyze the content of the variable, free of the problems that one can find in past work (incorrect model selection for small data-regimes, use of samples instead of variational parameters as inputs). We showed that there are particularly difficult datasets for which the first words are not very predictive of their labels, and therefore, these datasets should be systematically used in evaluations. Moreover, the agreement metric is another complementary evaluation that is automatic and focused on generation. We hope that these methods will see widespread adoption for measuring progress more reliably.

A promising research direction is to investigate the root cause behind memorization. A simple reason for the memorization of the first few words could be that, in the beginning of training, the reconstruction loss is higher on these words (see \textit{LSTM-LM} in Figures \ref{fig:loss_yahoo}, \ref{fig:loss_agnews}, \ref{fig:loss_amazon}, \ref{fig:loss_yelp}). These early errors should therefore account for a proportionally large part of the gradients and pressure the encoder to store information about the first words.
If that is correct, the left-to-right factorization of the decoder could be at fault, which would explain the successes of the unigram decoders. More powerful decoders with alternative factorizations could avoid this issue, for example, non-autoregressive Transformers \citep{gu_non-autoregressive_2017} or Transformers with flexible word orders \citep{gu_insertion-based_2019}.

VAEs operate on uncorrupted inputs and learn a corruption process in the latent space. In contrast, models in the BERT family \citep{devlin_bert_2018} are given corrupted inputs and are penalized only on these corrupted inputs, thereby avoid memorization altogether. Therefore, another research avenue would be to blend the two frameworks \citep{im_denoising_2017}.

\section*{Acknowledgements}
We thank NSERC for financial support, Calcul Canada for computational resources and Siva Reddy, Loren Lugosch, Makesh Narasimhan and Arjun Akula for their comments on the draft, as well as the reviewers and the meta-reviewer.

\bibliography{references,references_zotero}

\appendix

\clearpage
\section{On the use of KL annealing, the choice of the free bits flavor and resetting the decoder}
\label{appendix:modif_to_li}
\citet{li_surprisingly_2019} evaluated their models in the SSL setting (Section 3.3 of their paper). However, their experimental setting is not very rigorous. In the case of the 100 labeled examples, hyperparameter selection is done on a very large validation set of 10000 examples. However, the validation set here should be seen as nothing more than a split of the training data dedicated to optimising hyperparameters. In the words of \citet{cawley2010over}, ``model selection should be viewed as an integral part of the model fitting procedure''.
Besides methodological issues, we run our own hyperparameter search on the Yelp dataset to properly disentangle the effects of KL annealing, the free bits method and verify the importance of resetting the decoder. We use the semi-supervised learning setting presented in Section \ref{section_ssl} to evaluate the learned encoders.

\subsection{The free bits technique and variants}
\label{subsection_freebits}

The \textit{original} free bits objective \citep{kingma_improved_2016} is the following modification to the KL term: 

$$\sum_j^K \max(\frac{\lambda}{K}, KL(q_j(z_j|x)||p_j(z_j)))$$

where indices denote components. In this formulation, each component of the multivariate normal is allowed to deviate from the prior by a small amount. Instead, in the $\delta$-VAE formulation, one component can use of all the $\lambda$ free bits and the rest of the components can collapse to the prior. This is the variant called $\delta$, used throughout the paper: 

$$\max(\lambda, KL(q(z|x)||p(z)))$$

Other modifications of the free bits technique include the use of a variable coefficient in front of the KL term \citep{chen_variational_2016}, the target rate objective in \citet{alemi_fixing_2018}, minimum desired rate \citep{pelsmaeker_effective_2019}, etc. A comparison of all these methods is out of the scope of this paper and the $\delta$ variant satisfies our only requirement: the rate should be close to the desired rate. 

\subsection{KL annealing and the original free bits method higher the rate}

Our hypotheses are:
\begin{itemize}
	\item KL annealing aims at fixing the posterior collapse and is therefore redundant with the free bits,
	\item KL annealing performs this role by increasing capacity inconsistently across models, making them harder to compare,
	\item the original free bits formulation impose the unnecessary constraint that the free bits should be balanced over all components.
\end{itemize}

To study the influence of the free bits variant as well as of KL annealing, we use the same experimental protocol as described in Section \ref{section_ssl}. To save computations, we fix $d=16$. We do not perform model selection on the desired rate $\lambda$ in order to see which methods yield the rates that are closest to the desired rate. Table \ref{table_kl_annealing} shows these hypotheses are correct. Therefore, all the experiments in the paper use the $\delta$ variant without annealing. 

In \citet{li_surprisingly_2019}'s work, the original, per-component variant of the free bits might have been chosen because it trivially maximizes a metric called \textit{active units} (AU). However, to our knowledge, there is no evidence that this metric should be maximized, neither theoretical nor empirical. 

\begin{table*}[t]
\begin{center}
\begin{small}
\begin{sc}

\begin{tabular}{cc|ccccccc}
\toprule
       FB & $\lambda$ & Ann. &                  F1(5) &                 F1(50) &                F1(500) &               F1(5000) &          F1(all) &              KL \\
\midrule
        O &         2 &   10 &  $53.3\pm^{5.5}_{3.3}$ &  $69.8\pm^{1.8}_{1.3}$ &  $73.6\pm^{0.2}_{1.7}$ &  $74.0\pm^{0.1}_{1.8}$ &  $73.6\pm_{1.1}$ &   $5.27\pm_{0.47}$ \\
        O &         2 &    0 &  $51.8\pm^{4.8}_{6.7}$ &  $62.7\pm^{2.5}_{3.8}$ &  $67.0\pm^{0.4}_{5.6}$ &  $67.5\pm^{0.1}_{5.8}$ &  $66.9\pm_{2.7}$ &   $2.58\pm_{0.46}$ \\
 $\delta$ &         2 &   10 &  $51.7\pm^{4.6}_{4.7}$ &  $64.5\pm^{1.9}_{6.7}$ &  $68.3\pm^{0.4}_{7.3}$ &  $69.1\pm^{0.2}_{6.7}$ &  $68.4\pm_{3.3}$ &    $2.5\pm_{0.24}$ \\
 $\delta$ &         2 &    0 &  $58.7\pm^{5.5}_{3.2}$ &  $74.0\pm^{2.7}_{4.4}$ &  $78.1\pm^{0.3}_{4.1}$ &  $78.6\pm^{0.1}_{4.3}$ &  $78.6\pm_{1.9}$ &   $2.27\pm_{0.02}$ \\
 \midrule
        O &         8 &   10 &  $60.0\pm^{6.0}_{8.7}$ &  $77.5\pm^{1.2}_{2.2}$ &  $80.8\pm^{0.3}_{4.1}$ &  $81.2\pm^{0.1}_{4.2}$ &  $81.2\pm_{2.1}$ &  $10.67\pm_{0.44}$ \\
        O &         8 &    0 &  $60.2\pm^{7.3}_{4.7}$ &  $77.7\pm^{2.0}_{2.6}$ &  $81.4\pm^{0.3}_{2.2}$ &  $81.7\pm^{0.1}_{2.2}$ &  $81.5\pm_{0.9}$ &   $9.48\pm_{0.08}$ \\
 $\delta$ &         8 &   10 &  $57.6\pm^{7.6}_{4.2}$ &  $76.3\pm^{1.4}_{1.1}$ &  $80.3\pm^{0.3}_{3.0}$ &  $80.8\pm^{0.1}_{2.9}$ &  $80.3\pm_{1.0}$ &   $8.21\pm_{0.07}$ \\
 $\delta$ &         8 &    0 &  $60.4\pm^{4.1}_{3.6}$ &  $80.0\pm^{1.3}_{3.0}$ &  $82.7\pm^{1.0}_{0.9}$ &  $83.3\pm^{0.1}_{2.3}$ &  $83.5\pm_{0.8}$ &   $8.12\pm_{0.02}$ \\
\bottomrule
\end{tabular}

\end{sc}
\end{small}
\end{center}
\caption{$\delta$-VAE-style free bits with no KL annealing delivers the best SSL performance and the KL value closest to the desired rate. \textit{Ann.}: 0: no annealing, 10: anneal for 10 epochs; \textit{FB}: free bits type; \textit{F1($n$)}: F1-score in the $n$ data-regime; \textit{KL}: rate obtained after training.}
\label{table_kl_annealing}
\end{table*}

\subsection{On the importance of resetting the decoder after pretraining}

\citet{li_surprisingly_2019} proposed to pretrain an \textit{AE} with a reconstruction loss only. Then, the parameters of the decoder are re-initialised and the (modified) KL term is added to the objective. Since it is not very clear why it would be useful, we studied the impact of this choice. Table \ref{table_reset_dec} shows that it is is crucial. 

\begin{table*}[t]
\begin{center}
\begin{small}
\begin{sc}

\begin{tabular}{cc|cccccc}
\toprule
Reset. & $\lambda$ &                  F1(5) &                 F1(50) &                F1(500) &               F1(5000) &          F1(all) &             KL \\
\midrule
        N &         2 &  $51.0\pm^{4.2}_{5.6}$ &  $61.3\pm^{2.0}_{9.2}$ &  $65.6\pm^{0.5}_{9.2}$ &  $66.2\pm^{0.1}_{9.5}$ &  $65.2\pm_{4.9}$ &  $2.36\pm_{0.15}$ \\
        Y &         2 &  $58.7\pm^{5.5}_{3.2}$ &  $74.0\pm^{2.7}_{4.4}$ &  $78.1\pm^{0.3}_{4.1}$ &  $78.6\pm^{0.1}_{4.3}$ &  $78.6\pm_{1.9}$ &  $2.27\pm_{0.02}$ \\
        N &         8 &  $57.4\pm^{5.6}_{2.4}$ &  $73.4\pm^{1.5}_{7.3}$ &  $77.2\pm^{0.3}_{6.6}$ &  $77.5\pm^{0.1}_{6.7}$ &  $77.4\pm_{2.6}$ &  $8.23\pm_{0.08}$ \\
        Y &         8 &  $60.4\pm^{4.1}_{3.6}$ &  $80.0\pm^{1.3}_{3.0}$ &  $82.7\pm^{1.0}_{0.9}$ &  $83.3\pm^{0.1}_{2.3}$ &  $83.5\pm_{0.8}$ &  $8.12\pm_{0.02}$ \\
\bottomrule
\end{tabular}

\end{sc}
\end{small}
\end{center}
\caption{Resetting the decoder brings very noticeable gains on all data-regimes and with different rates. Yelp dataset, $\delta$-VAE free bits, no KL annealing. For columns interpretations, see Table \ref{table_kl_annealing}.}
\label{table_reset_dec}
\end{table*}

\section{Further evidence for memorization}
\label{app_memorization}

\begin{table}[t]
\begin{adjustbox}{max width=0.48\textwidth}
\begin{tabular}{l|lllll}
\toprule
Dataset & Splits size & Label & $|\mathcal{Y}|$ & $H[Y]$ & NLL\\
\midrule
	AGNews & 110/10/10 & Topic & 4 & 1.39 & $128.77${\small $\pm0.21$}\\
	Amazon & 100/10/10 & Sent. & 5 & 1.61 & $82.90${\small $\pm0.10$}\\
	Yahoo & 100/10/10 & Topic &  10 & 2.30 & $81.91${\small $\pm0.36$}\\
	Yelp & 100/10/10 & Sent. & 2 & 0.67& $34.60${\small $\pm0.28$}\\
\bottomrule
\end{tabular}
\end{adjustbox}
\caption{Datasets characteristics. $|\mathcal{Y}|$: number of different labels. $H[Y]$: entropy of labels. NLL: mean negative log-likelihood of LSTM baseline models (std. over 3 runs). Splits size: train/valid/test sizes in thousands.}
\label{table_datasets}
\end{table}

\begin{figure*}
  \includegraphics[scale=0.6]{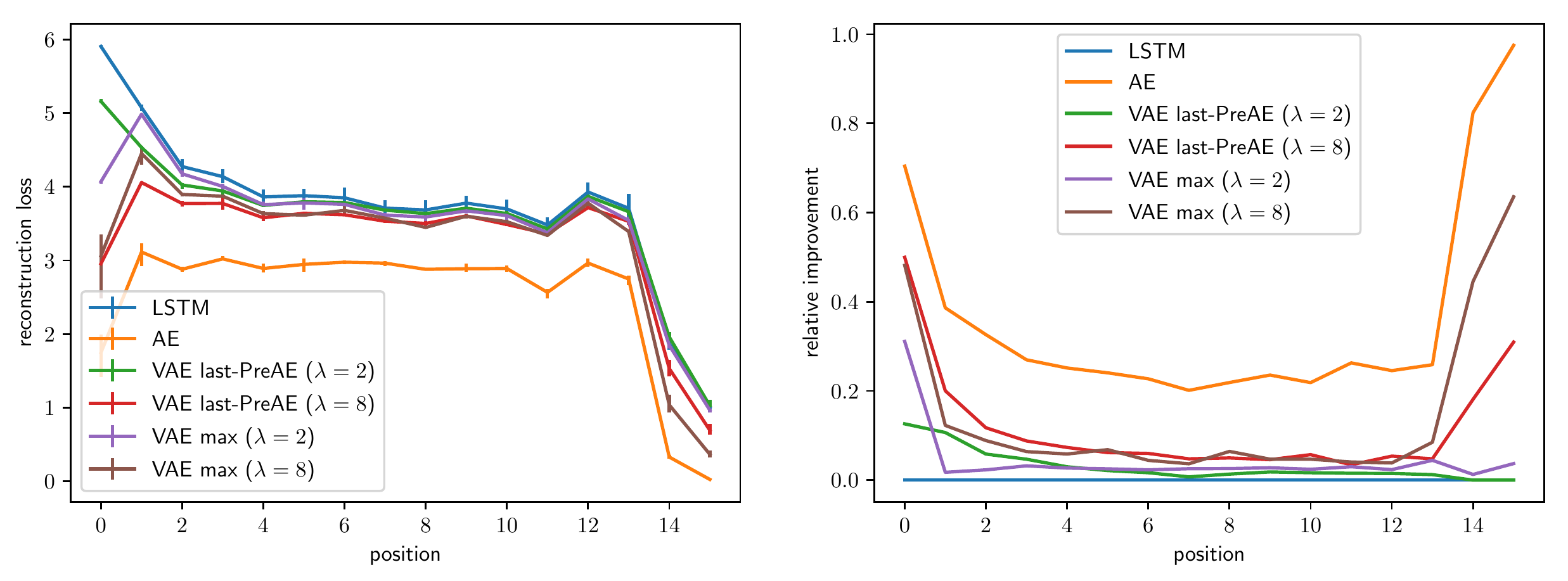}
  \caption{Reconstruction loss as a function of word position on the AGNews dataset. See Figure \ref{fig:loss_yahoo}.}
	\label{fig:loss_agnews}
\end{figure*}

\begin{figure*}
  \includegraphics[scale=0.6]{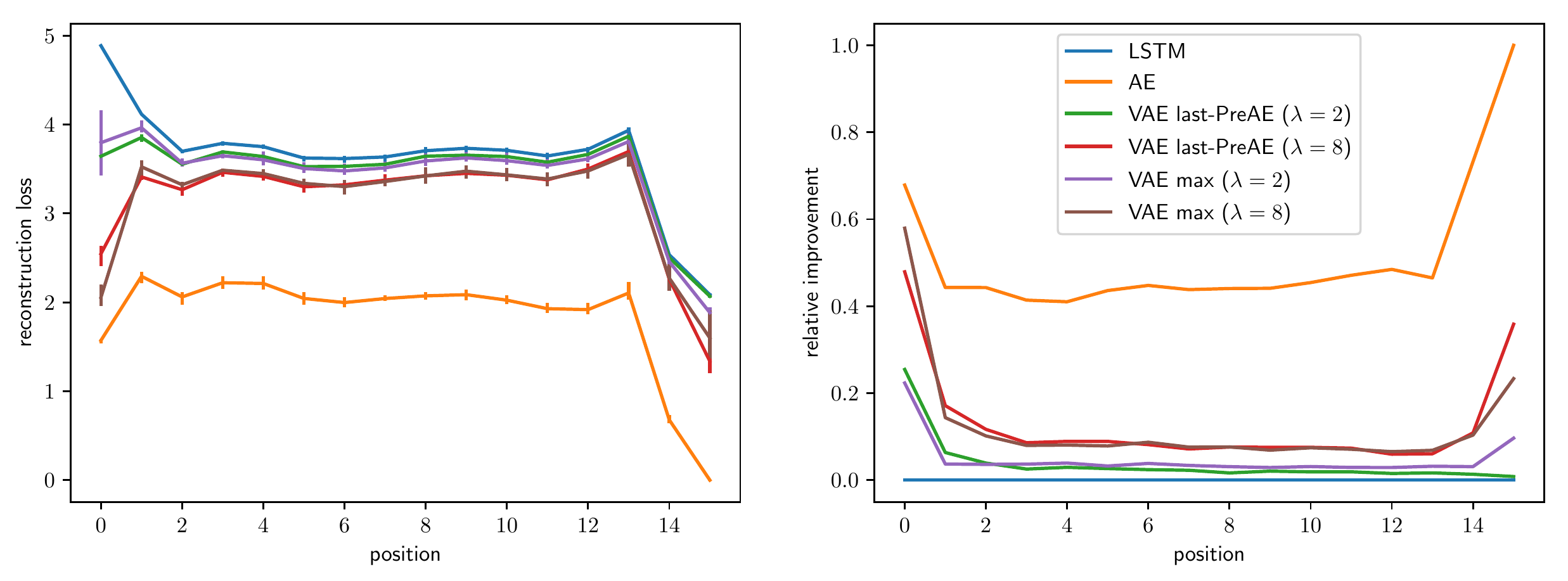}
        \caption{Reconstruction loss as a function of word position on the Amazon dataset. See Figure \ref{fig:loss_yahoo}.}
	\label{fig:loss_amazon}
\end{figure*}

\begin{figure*}
  \includegraphics[scale=0.6]{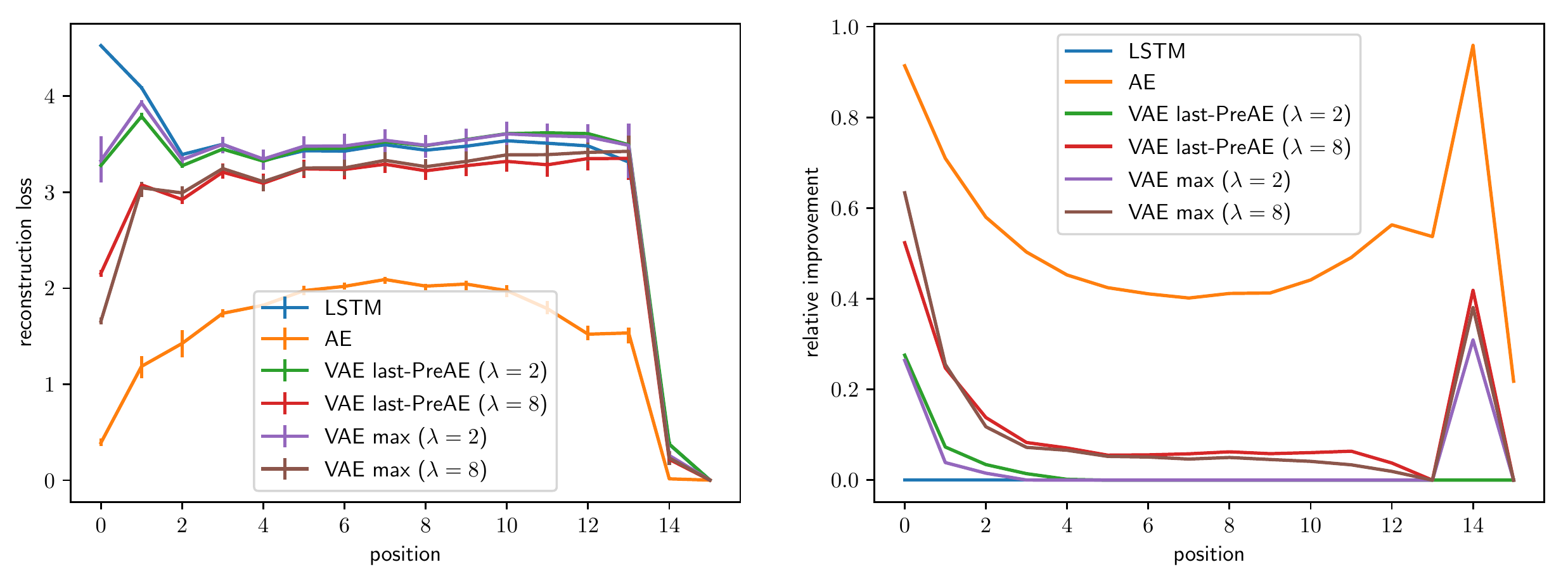}
        \caption{Reconstruction loss as a function of word position on the Yelp dataset. See Figure \ref{fig:loss_yahoo}.}
	\label{fig:loss_yelp}
\end{figure*}

\subsection{Plots on other datasets}
\label{app_other_plots}

Figures \ref{fig:loss_agnews}, \ref{fig:loss_amazon}, and \ref{fig:loss_yelp} show the reconstruction loss and the relative improvement on other datasets. 

On the Yelp dataset, the penultimate token is a punctuation mark which is always followed by the end-of-sentence token, so predicting its position is equivalent to predicting the sentence length. That is why the peak at the end occurs before the last token. Moreover, on Yelp, the situation is worse with $\lambda=2$: between positions 6 and 13, not only is there no improvement, but the reconstruction is higher than that of the baseline.

\subsection{Tracing back reconstruction gains to words}

\label{app_tracing_back} 

If words in a document were independently modeled, any improvement in reconstruction at a certain position would indicate that information about the word in that position were encoded in the latent variable. However, words are far from being independently predicted, so how can we trace back the information to the encoder? 

First, any latent information related to the first word should not yield any improvements on the prediction of the second word, because the decoder is recurrent and trained using teacher forcing, i.e., conditioned on the true first word, so that information would be redundant. However, information related to the second word in the latent variable can help the decoder predict the first word. Therefore, gains in position $i$ can only be attributed to information pertaining to the words in positions $\geq i$. 

Second, the correlation between words in two positions decreases as the distance between these words grow. In effect, information pertaining to the second word yields more gains on the second word than on the first word. From these two facts, we conclude that gains for a position $i$ mostly comes from information about the word in position $i$ itself. 

\subsection{Reconstruction and memorization}

\label{app_decoding} 

To study the concrete impact of this observation for generation, we encode and decode test documents using the \textit{last-PreAE} variant.\footnote{$\lambda=8$, $d=16$, beam search with beam size of 5.} Then, we compute the ratio of documents for which the first word in the sources and in the reconstructions match and similarly, how often the sources and their reconstructions have the same number of words. 

We compare these with scores obtained by a baseline model that outputs the most frequent first word given the label and the most common document length given the label. This baseline mimicks the behavior of a hypothetical VAE which would encode the labels of the documents (topic or sentiment) perfectly and nothing more.

Results in Table \ref{table_reconstruction_acc} show that with the \textit{last-PreAE} the first words are reconstructed with much higher accuracy than if the latent vector only encoded the label. On the last two datasets, it recovers the first words on more than half of the documents whereas the baseline only recovers the first words between 11.3 and 14.1\% of the time. Accurate encoding of the number of words seems less systematic than the encoding of the first few words. For example, on AGNews, the sentence length is recovered less often than our baselines. The encoding of the sentence length is more pronounced on datasets with small documents like Yahoo and Yelp.

\begin{table}[t]
\begin{center}
\begin{adjustbox}{max width=0.48\textwidth}
\begin{tabular}{c|cc|cc}
\toprule
 & \multicolumn{2}{c|}{\textit{last-PreAE}} & \multicolumn{2}{c}{Clf. given label} \\
Dataset & 1st (\%) & Len. (\%) & 1st (\%) & Len. (\%) \\
\midrule
AGNews & $29.6${\small $\pm1.1$} &  $3.6${\small $\pm0.1$} &   $12.9$ & $4.8$ \\
Amazon &  $42.4${\small $\pm2.3$} &  $13.0${\small $\pm1.6$} & $14.0$ & $0  $ \\
Yahoo &  $56.6${\small $\pm1.0$} &  $17.1${\small $\pm1.1$} &  $11.3$ & $4.9$ \\
Yelp &  $53.0${\small $\pm0.5$} &  $33.7${\small $\pm1.7$} &   $14.1$ & $9.7$ \\
\bottomrule
\end{tabular}
\end{adjustbox}
\end{center}
\caption{The latent variables encode more information than the label alone, in particular, information that allows to retrieve the first word and the document length with high accuracy.}
\label{table_reconstruction_acc}
\end{table}

\section{Training procedure}

\label{appendix:training}
\subsection{Grid search}

The target rates $\lambda$ are chosen to be higher than the entropy of the labels of the documents (Table \ref{table_datasets}) as we assume that the latent variable should at least capture the annotated label. Indeed, $\lambda=2$ nats is enough to store the labels of all datasets without any loss, except Yahoo which has an entropy of $2.3$ whereas $\lambda=8$ nats suffices to capture much more information than needed to store the labels on all datasets. Moreover, these rates are chosen to be much smaller than the reconstruction loss of the baselines because of the technical difficulty of increasing the rate without degrading the log-likelihood explained above.

The latent vector dimension $d$ is either 4 or 16. Recall that our representations are evaluated on downstream tasks with very limited data in some cases (as little as 5 examples per class), so we need a small enough dimension of latent vector to be able to learn. We suppose that $d=4$ will be favored for the 5 or 50 examples per class regime while $d=16$ could be more efficient above this, but we leave this choice to the model selection procedure.

\subsection{Constant hyperparameters}

All the runs are trained using SGD with a learning rate of $0.5$ and gradients are clipped when their norms are higher than 5. We use the following early stopping scheme: at every epoch, if there has not been improvements on the validation error for two epochs in a row, the learning rate is halved. Once it has been halved four times, the training stops.

All the LSTMs have hidden state size of 512 and use a batch size of 64. No dropout is applied to the encoders. The LSTM decoders use dropout ($p=0.5$) both on embeddings and on the hidden states (before the linear transformation that gives logits). Similarly, dropout is applied to the representation before the linear transformation that gives the logits for the Unigram decoder. Word embeddings are initialized randomly and learned.

\subsection{Computing infrastructure and average runtime}

We performed training and evaluations of the models on a cluster containing a hundred of GPUs with various specifications (NVIDIA Tesla k80, Titan X, Titan Xp, etc.). Given that all the datasets have roughly 100000 training examples (cf. Table \ref{table_datasets}) and that neural networks are trained with BPTT \citep{werbos1990backpropagation}, the training time mostly depends on the average sentence length and the vocabulary size. Pretraining schemes (\textit{PreAE}, \textit{PreUni} and \textit{PreLM}) require the training of two models. Roughly, the training time of a single model (pretraining or final) varied between 1 hour and 6 hours.

To be more specific, the best baselines and our best variants use pretraining phases (\textit{PreAE} from \citet{li_surprisingly_2019} and \textit{PreUni}, respectively). \textit{PreUni} is faster because the first training phase uses a non-recurrent decoder, and the second training phase does not backpropagate and does not update the encoder. However, \textit{PreAE} does not require pretraining for each desired target rate $\lambda$, unlike our approach. Overall, the approaches have comparable runtimes. 

Some of our models offer an interesting compromise: \textit{BoW-max-LSTM} with no pretraining and a simpler architecture is probably the fastest, yet outperform the \textit{PreAE} baselines.

\section{Related work}
\subsection{Related models}
\label{section_related_models}
The models that we use are similar to already proposed models.

The NVDM model of \citet{miao_neural_2016} is precisely \textit{BoW-max-Uni}.

\citet{zhao_learning_2017} proposed to use two reconstruction losses: the regular reconstruction loss given by the recurrent decoder and an auxiliary loss computed from a unigram decoder. In comparison, our \textit{Uni} models are trained in two steps: the encoder is trained jointly with the unigram decoder, then the decoder is thrown away and we train a recurrent decoder using the fixed encoder. This way, one decoder cannot dominate the other and we do not have to deal with an additional hyperparameter to weight the two losses. 

Instead of using an auxiliary \textit{loss}, we have an auxiliary \textit{decoder} that is only used for the purpose of training the encoder. This method was presented by \citet{de_fauw_hierarchical_2019} for training generative models of image. There is a slight difference: they use a feedforward auxiliary decoder to produce different probability distributions for all the pixels, whereas our unigram probability distribution is the same for all words of a document. This modification allows us to deal with varying lengths of documents.

Finally, the \textit{PreLM} training procedure is related to large LM pretraining in the spirit of contextualized embeddings \citep{peters_deep_2018} and its successors. Note, however, two differences. Firstly, we do not use external data and stick to each individual training set, because the goal is not to evaluate transfer learning abilities. Secondly, we do not fine-tune the entire encoder, but only learn the linear transformations $L_1$ and $L_2$ that produce the variational parameters, to make sure that the VAE objective will have no impact on the extraction of features. 

\subsection{Methods and evaluations}

\label{sec:rel_eval}

In their analysis of the semi-amortized VAE, \citet{kim_semi_2018} use several saliency measures (defined as expectations of gradients) to determine which words influence the latent variable, or are influenced by it. Using these measures, they noticed that the beginning of the sentence and the end-of-sentence token have a large influence on the variable. Our method is very similar, but slightly simpler and directly interpretable in terms of quantity of information (in nats). 

\citet{ficler_controlling_2017} learn \textit{LSTM-LMs} conditioned on labels that describe high-level properties of texts. Among others, they want to verify that generated texts exhibit the same properties as the conditioning labels. For instance, when the \textit{LSTM-LM} is conditioned on positive sentiment value, the generated texts should also exhibit a positive sentiment. To check that the conditioning variables and the generated texts are consistent, they use the following procedure. First, they extract information about the various documents using heuristics or with the help of annotators. Then, they learn \textit{LSTM-LM}s conditioned on these labels. Finally, they quantify the ratio of generated samples which have the same labels than the conditioning labels, either by applying the same heuristics again to the generated samples or by asking human annotators once more. Our evaluation in Section \ref{section_text_gen} is similar; we simply replace the heuristics and the human annotators with classifiers learned on ground-truth data. 

\citet{mccoy_rnns_2019} trained autoencoders with different combinations of encoders and decoders (unidirectional, bidirectional or tree-structured) and decomposed the representations learned by the encoders using tensor product representations \citep{smolensky_tensor_1990}. They find that decoders ``largely dictate'' the way information is encoded. This is in line with our own conclusions. An important difference between our works is that they study how information is encoded in sequence-to-sequence models without capacity limitations, whereas in our study, the VAE objective puts severe constraints on the capacity.

\section{Semi-supervised learning experiments}

\subsection{Model selection}
\label{app_cv}

For a given dataset in a given data-regime, we want a measure of the performance of our models that abstracts away from i) hyperparameters for the VAEs, ii) hyperparameters for the downstream task classifiers, iii) subsampling of the dataset and iv) parameter initialisation of the VAEs. As is usually done by practitioners, we optimize over the hyperparameters of the VAEs and the classifiers, eliminating i) and ii) as sources of variance. We can study the robustness of the models by looking at the variance induced by the choice of the subsample and the initialisation of the parameters. 

On a given dataset and in a given data-regime, for a given model, we note $F_{ij}^{H_M,H_C}$ the F1-score obtained on the test set on the subsample using seed $i$, the parameter initialisation using seed $j$, VAE hyperparameters $H_M$ and classifier hyperparameters $H_C$. We use repeated stratified K-fold cross-validation \citep{moss_using_2018} to compute a validation error $\widehat{F_{ij}^{H_M,H_C}}$. For all training folds, we train logistic regression classifiers with $L_2$ regularisation and a grid-search on $H_C \in \{0.01, 0.1, 1, 10, 100\}$. We select the best classifier hyperparameter: 

$$H_C^* = \argmax_{H_C} \widehat{F_{ij}^{H_M,H_C}}$$

Then, the best VAE hyperparameter is chosen by averaging over the $s=3$ random seeds and picking the best classifier hyperparameter, 

$$H_M^* = \argmax_{H_M} \frac{1}{s} \sum_{i=1}^s \widehat{F_{ij}^{H_M, H_C^*}}$$

Having optimised the hyperparameters, we compute the test set F1-score:

$$F_{ij} = F_{ij}^{H_M^*, H_C^*}$$

\subsection{Decomposing the variances of the scores}
\label{appendix:stats_ssl}

For a given model, dataset and data-regime, after optimisation of the hyperparameters of the VAE and the classifier, we collect several F1-scores $F_{ij}$ which depend on the seed used to subsample the dataset $i$ and the seed used to initialise the model parameters $j$. We posit a linear model with one random-effect factor, the initialisation seed, and where replicates are obtained by varying the subsampling seed: 
$$F_{ij} = \mu + \alpha_j + \eps_{ij}$$
Assuming that $\alpha_j$ and $\eps_{ij}$ are independent random variables with null expectations, we can decompose the variance as 
\begin{align*}
    \Var(F_{ij}) &= \E[ (F_{ij} - \mu)^2 ] \\
               &= \E[ (\alpha_i + \eps_{ij}) ^2 ] \\
               &= \E [\alpha_i ^ 2] + \E[\eps_{ij}^2] \\
               &= \Var(\alpha_i) + \Var(\eps_{ij}) 
\end{align*}
This is the basis of the method of analysis of variance (ANOVA) and is often used to test hypotheses (for instance, that the effect $\E[\alpha_i]$ is significant) \citep{oehlert_first_2010}. The two estimates of $\sigma_{init}^2$ and $\sigma^2$ are usually denoted $MS_T$ and $MS_E$.

In our case, we are only interested in estimating roughly what variability is due to the model initialisation and what is due to the subsampling of the dataset. 

Note that we could treat the two sources of variance $i$ and $j$ symmetrically by adding add a term $\beta_i$, but we would need to report 3 standard deviations (that of $\alpha_j$, $\beta_i$ and $\eps_{ij}$) to get the full picture. The most important estimate is $\sigma_{init}$. It quantifies the inherent robustness of the model to different initialisations. The effect of the subsampling is specific to the dataset, therefore, it is less relevant to our analysis.

\subsection{What is the representation of a document?}
\label{section_mean_vs_samples}

VAEs are mostly used for generating samples but are also sometimes used as feature extractors for SSL. In the latter case, it is not clear what the representation of a datapoint is: the mean of the approximate posterior $\mu$ or the noisy samples $Z \sim \mathcal{N}(\mu, I \sigma^2)$? \citet{kingma_semi-supervised_2014} feed noisy samples $z$ in the classifiers but in the literature of VAEs applied to language modeling, it is more common to use $\mu$ without explanation or even mention.\footnote{For instance, \citet{li_surprisingly_2019} and \citet{fu_cyclical_2019} do not mention what representation they use but their code uses the mean; \citet{long_preventing_2019} report using a concatenation of the mean and the variance vectors.}

If we are interested purely in downstream task performance, the mean should perform best, as the samples are just noisy versions of the mean vector (it is still not completely straightforward as the noise could play a regularizing role). However, in order to evaluate what information is \textit{effectively} transmitted to the decoder, we should use the samples. The performance of downstream task classifiers using the mean does not tell us \textit{at all} whether the latent variable is used by the decoder to reconstruct the input. The following experiment illustrates this fact.

We train the original VAE architecture on the Yelp dataset, both with and without the \textit{PreAE}, using the original ELBo objective ($\lambda=0$). As expected, the KL term collapses to 0. Then, we train a classifier using the procedure explained above using 5000 examples per class. We expect that its performance will be close to random chance, regardless of whether samples or the mean parameter are used as inputs. However, Table \ref{table_mu_vs_z} shows that this is not the case. Using samples, we do get random chance predictions from the classifiers, whereas using means, the performance is remarkably high (as high as 81.5 of F1 using pretraining). The reason is that the KL term never \textit{completely} collapses to 0. Therefore, $\mu$ can be almost zero while still encoding a lot of information about its inputs. However, when the KL term is close to 0, the variance of the samples is close to 1, so no information is transmitted to the decoder. This tendency is exacerbated with the \textit{PreAE} runs, for which the means encode remnants of the pretraining phase.

\begin{table}[t]
\begin{center}
\begin{adjustbox}{max width=0.40\textwidth}
\begin{tabular}{c|c|c|c}
\toprule
\multirow{2}{*}{PreAE} & \multicolumn{2}{c|}{F1} & \multirow{2}{*}{KL} \\
~ & $z$ & $\mu$ & ~ \\

\midrule
             No & 49.5 & 64.7 &     $1 e^{-4}$ \\
             Yes & 49.6 & 81.5 &   $2 e^{-4}$ \\
\bottomrule
\end{tabular}
\end{adjustbox}
\end{center}
\caption{When the KL collapses, the performances of classifiers trained on the mean $\mu$ vs on samples $z \sim \mathcal{N}(\mu, I \sigma^2)$ are very different, especially for pretrained models. $z$ does not contain any information while $\mu$ is very predictive of the label.} 
\label{table_mu_vs_z}

\end{table}

This experiment shows that it is crucial to report what representation ($z$ or $\mu$) is analyzed and to cautiously interpret the results. Therefore, for the purpose of analysing representations for text generation, we feed $z$ as inputs to the classifiers.

\begin{table*}[t]

\begin{center}
\begin{small}
\begin{adjustbox}{max width=0.84\textwidth}
\begin{tabular}{l|Hccc|cHcccHc}
\toprule                  
 & & & & &                5 &             10 &             50 &            500 &           5000 & All & All   \\
 & Enc. & $r$ & Dec. & Pre. &                \multicolumn{7}{c}{F1$\pm^{\sigma}_{\sigma_{\mathrm{init}}}$} \\
 \midrule
 \multirow{9}{*}{\rotatebox[origin=c]{90}{AGNews}} &   LSTM &  last &  LSTM &      - &  $59.6\pm^{5.1}_{11.9}$ &  $66.6\pm^{2.5}_{14.1}$ &  $71.7\pm^{1.0}_{12.1}$ &  $73.6\pm^{0.1}_{11.8}$ &  $73.7\pm^{0.1}_{11.9}$ &   $73.3\pm^{-}_{5.4}$ &   $73.6\pm^{-}_{5.4}$ \\
 & LSTM &  last &  LSTM &  AE &   $65.8\pm^{3.3}_{3.3}$ &   $73.9\pm^{4.2}_{6.6}$ &   $81.0\pm^{0.7}_{1.1}$ &   $82.8\pm^{0.3}_{0.6}$ &   $83.1\pm^{0.1}_{0.7}$ &   $82.7\pm^{-}_{0.4}$ &   $83.4\pm^{-}_{0.3}$ \\
 & LSTM &   max &  LSTM &      - &   $27.3\pm^{2.4}_{1.2}$ &   $30.2\pm^{2.1}_{6.6}$ &   $30.8\pm^{3.4}_{5.4}$ &  $33.1\pm^{0.9}_{10.5}$ &   $33.8\pm^{0.4}_{8.6}$ &   $31.0\pm^{-}_{3.1}$ &   $34.6\pm^{-}_{2.4}$ \\
 & LSTM &   max &  LSTM &  AE &  $55.7\pm^{4.5}_{18.7}$ &  $64.8\pm^{6.0}_{15.5}$ &   $75.1\pm^{1.3}_{2.6}$ &   $81.9\pm^{0.3}_{0.0}$ &   $82.5\pm^{0.1}_{0.4}$ &   $82.2\pm^{-}_{0.4}$ &   $83.3\pm^{-}_{0.4}$ \\
 & BoW &   max &  LSTM &      - &   $72.7\pm^{2.0}_{5.9}$ &    $77.1\pm^{2.2}_{3.2}$ &   $81.2\pm^{0.6}_{0.8}$ &   $82.2\pm^{0.2}_{0.8}$ &   $82.3\pm^{0.1}_{1.0}$ &   $82.0\pm^{-}_{0.3}$ &   $83.1\pm^{-}_{0.3}$ \\
 &LSTM &   max &   Uni &      - &   $71.6\pm^{5.5}_{0.1}$ &   $75.0\pm^{4.3}_{1.6}$ &   $80.4\pm^{0.8}_{0.7}$ &   $81.8\pm^{0.5}_{0.5}$ &   $82.4\pm^{0.1}_{0.4}$ &   $82.2\pm^{-}_{0.4}$ &   $83.9\pm^{-}_{0.3}$ \\
 &LSTM &  last &   Uni &      - &  $54.8\pm^{5.2}_{57.1}$ &  $58.2\pm^{3.3}_{65.6}$ &  $61.7\pm^{0.8}_{71.4}$ &  $62.9\pm^{0.4}_{71.0}$ &  $63.0\pm^{0.3}_{71.1}$ &  $61.5\pm^{-}_{34.5}$ &  $59.3\pm^{-}_{40.9}$ \\
 & BoW &   max &   Uni &      - &   $71.8\pm^{4.5}_{1.8}$ &   $75.6\pm^{2.6}_{1.6}$ &   $81.4\pm^{0.5}_{0.6}$ &   $82.5\pm^{0.1}_{0.5}$ &   $82.5\pm^{0.1}_{0.6}$ &   $82.3\pm^{-}_{0.3}$ &   $83.1\pm^{-}_{0.5}$ \\
 &LSTM &   avg &  LSTM &      LM &   $70.8\pm^{4.8}_{4.3}$ &   $76.3\pm^{3.0}_{2.7}$ &   $81.2\pm^{0.9}_{1.2}$ &   $82.6\pm^{0.2}_{1.3}$ &   $82.8\pm^{0.1}_{0.9}$ &   $82.5\pm^{-}_{0.4}$ &   $83.5\pm^{-}_{0.1}$ \\

\midrule
 \multirow{9}{*}{\rotatebox[origin=c]{90}{Amazon}} & LSTM &  last &  LSTM &      - &  $18.9\pm^{1.7}_{0.5}$ &  $19.3\pm^{1.2}_{0.3}$ &  $20.9\pm^{1.2}_{0.9}$ &  $22.5\pm^{0.7}_{0.7}$ &  $23.3\pm^{0.4}_{1.1}$ &  $23.2\pm^{-}_{0.6}$ &  $22.9\pm^{-}_{1.5}$ \\
 & LSTM &  last &  LSTM &  AE &  $20.0\pm^{2.2}_{0.9}$ &  $21.2\pm^{1.4}_{1.5}$ &  $24.7\pm^{0.7}_{2.8}$ &  $27.2\pm^{0.4}_{3.1}$ &  $27.7\pm^{0.3}_{3.8}$ &  $28.1\pm^{-}_{1.1}$ &  $28.1\pm^{-}_{1.0}$ \\
 & LSTM &   max &  LSTM &      - &  $19.8\pm^{0.7}_{0.5}$ &  $19.8\pm^{1.1}_{0.5}$ &  $20.4\pm^{1.1}_{0.9}$ &  $22.2\pm^{0.6}_{2.1}$ &  $23.0\pm^{0.3}_{1.9}$ &  $22.6\pm^{-}_{0.7}$ &  $23.7\pm^{-}_{0.5}$ \\
 & LSTM &   max &  LSTM &  AE &  $22.3\pm^{2.6}_{0.7}$ &  $24.9\pm^{3.8}_{1.5}$ &  $30.5\pm^{0.9}_{3.0}$ &  $33.4\pm^{0.4}_{4.1}$ &  $34.1\pm^{0.3}_{4.8}$ &  $34.0\pm^{-}_{1.6}$ &  $34.0\pm^{-}_{1.6}$ \\
 &  BoW &   max &  LSTM &      - &  $21.0\pm^{2.6}_{1.1}$ &  $27.6\pm^{3.0}_{1.2}$ &  $34.6\pm^{0.7}_{1.1}$ &  $38.3\pm^{0.4}_{1.0}$ &  $39.0\pm^{0.1}_{0.6}$ &  $39.2\pm^{-}_{0.6}$ &  $38.9\pm^{-}_{0.7}$ \\
 & LSTM &   max &   Uni &      - &  $21.8\pm^{3.1}_{1.6}$ &  $25.9\pm^{3.1}_{1.3}$ &  $32.8\pm^{0.8}_{1.7}$ &  $36.9\pm^{0.4}_{0.9}$ &  $38.0\pm^{0.2}_{0.6}$ &  $36.8\pm^{-}_{0.7}$ &  $38.2\pm^{-}_{0.5}$ \\
 & LSTM &  last &   Uni &      - &  $24.0\pm^{3.0}_{1.0}$ &  $28.2\pm^{1.9}_{3.3}$ &  $31.2\pm^{0.6}_{1.4}$ &  $35.1\pm^{0.4}_{2.2}$ &  $36.1\pm^{0.2}_{2.4}$ &  $36.3\pm^{-}_{1.1}$ &  $36.8\pm^{-}_{0.9}$ \\
 & BoW &   max &   Uni &      - &  $25.4\pm^{3.2}_{0.2}$ &  $29.0\pm^{2.4}_{2.8}$ &  $32.8\pm^{1.0}_{1.3}$ &  $36.1\pm^{0.4}_{0.7}$ &  $36.9\pm^{0.2}_{0.8}$ &  $37.4\pm^{-}_{0.2}$ &  $37.9\pm^{-}_{0.2}$ \\
 & LSTM &   avg &  LSTM &      LM &  $21.8\pm^{3.8}_{0.6}$ &  $28.9\pm^{1.8}_{1.0}$ &  $35.3\pm^{0.8}_{0.4}$ &  $40.2\pm^{0.4}_{0.4}$ &  $41.1\pm^{0.2}_{0.4}$ &  $41.1\pm^{-}_{0.8}$ &  $40.0\pm^{-}_{0.4}$ \\

\midrule
 \multirow{9}{*}{\rotatebox[origin=c]{90}{Yahoo}} &LSTM &  last &  LSTM &      - &  $10.9\pm^{0.9}_{0.5}$ &  $10.6\pm^{1.3}_{0.5}$ &  $12.1\pm^{0.6}_{0.6}$ &  $13.9\pm^{0.4}_{2.1}$ &  $14.1\pm^{0.2}_{2.8}$ &  $14.2\pm^{-}_{1.4}$ &  $14.9\pm^{-}_{1.0}$ \\
 &LSTM &  last &  LSTM &  AE &  $20.7\pm^{0.7}_{0.5}$ &  $23.1\pm^{1.6}_{1.6}$ &  $32.2\pm^{0.8}_{0.6}$ &  $36.1\pm^{0.2}_{0.1}$ &  $36.7\pm^{0.1}_{0.5}$ &  $36.5\pm^{-}_{0.1}$ &  $37.2\pm^{-}_{0.7}$ \\
 &LSTM &   max &  LSTM &      - &   $9.9\pm^{1.0}_{1.3}$ &  $11.1\pm^{0.8}_{1.3}$ &  $13.0\pm^{0.6}_{2.1}$ &  $14.6\pm^{0.3}_{2.8}$ &  $14.9\pm^{0.1}_{3.1}$ &  $14.8\pm^{-}_{1.5}$ &  $15.7\pm^{-}_{0.5}$ \\
 &LSTM &   max &  LSTM &  AE &  $20.8\pm^{1.3}_{2.3}$ &  $23.2\pm^{1.6}_{1.2}$ &  $31.3\pm^{0.7}_{1.4}$ &  $35.6\pm^{0.3}_{1.2}$ &  $36.3\pm^{0.1}_{1.1}$ &  $36.0\pm^{-}_{0.7}$ &  $36.6\pm^{-}_{0.7}$ \\
 & BoW &   max &  LSTM &      - &  $23.4\pm^{2.1}_{2.9}$ &  $26.2\pm^{2.1}_{1.2}$ &  $36.7\pm^{1.1}_{0.5}$ &  $41.1\pm^{0.2}_{0.8}$ &  $41.6\pm^{0.1}_{0.9}$ &  $41.6\pm^{-}_{0.3}$ &  $42.6\pm^{-}_{0.2}$ \\
 &LSTM &   max &   Uni &      - &  $24.9\pm^{1.3}_{2.2}$ &  $25.9\pm^{2.5}_{0.7}$ &  $33.2\pm^{0.7}_{3.6}$ &  $37.3\pm^{0.1}_{3.1}$ &  $37.9\pm^{0.1}_{3.1}$ &  $37.6\pm^{-}_{1.4}$ &  $38.9\pm^{-}_{1.7}$ \\
 &LSTM &  last &   Uni &      - &  $24.5\pm^{3.8}_{1.7}$ &  $27.6\pm^{2.2}_{1.5}$ &  $30.8\pm^{1.7}_{0.6}$ &  $34.4\pm^{0.3}_{5.0}$ &  $35.1\pm^{0.1}_{4.7}$ &  $35.1\pm^{-}_{1.9}$ &  $37.1\pm^{-}_{2.3}$ \\
 & BoW &   max &   Uni &      - &  $24.1\pm^{2.9}_{2.7}$ &  $26.0\pm^{5.0}_{1.8}$ &  $35.0\pm^{0.9}_{1.2}$ &  $39.1\pm^{0.1}_{1.8}$ &  $39.5\pm^{0.1}_{1.7}$ &  $39.2\pm^{-}_{0.4}$ &  $40.1\pm^{-}_{0.7}$ \\
 &LSTM &   avg &  LSTM &      LM &  $21.9\pm^{2.3}_{1.3}$ &  $26.1\pm^{2.0}_{1.1}$ &  $36.1\pm^{0.8}_{0.7}$ &  $39.9\pm^{0.2}_{0.6}$ &  $40.4\pm^{0.1}_{0.4}$ &  $40.9\pm^{-}_{0.4}$ &  $41.7\pm^{-}_{0.3}$ \\

\midrule
 \multirow{9}{*}{\rotatebox[origin=c]{90}{Yelp}} & LSTM &  last &  LSTM &      - &    $49.9\pm^{4.5}_{2.7}$ &   $52.2\pm^{3.7}_{1.4}$ &  $55.6\pm^{2.3}_{2.9}$ &  $57.9\pm^{1.1}_{2.5}$ &  $59.5\pm^{0.2}_{2.7}$ &  $55.5\pm^{-}_{1.8}$ &  $61.9\pm^{-}_{2.5}$ \\
 & LSTM &  last &  LSTM &  AE &    $59.3\pm^{5.4}_{2.9}$ &   $65.6\pm^{4.9}_{3.2}$ &  $80.0\pm^{1.3}_{3.0}$ &  $82.7\pm^{1.0}_{0.9}$ &  $83.3\pm^{0.1}_{2.3}$ &  $83.5\pm^{-}_{0.8}$ &  $67.9\pm^{-}_{0.1}$ \\
 & LSTM &   max &  LSTM &      - &    $61.6\pm^{8.2}_{8.8}$ &   $62.1\pm^{6.5}_{5.3}$ &  $71.4\pm^{2.3}_{6.3}$ &  $76.0\pm^{0.2}_{2.3}$ &  $76.5\pm^{0.1}_{2.0}$ &  $76.4\pm^{-}_{0.5}$ &  $78.0\pm^{-}_{1.7}$ \\
 & LSTM &   max &  LSTM &  AE &   $59.9\pm^{10.4}_{7.9}$ &   $62.9\pm^{4.4}_{2.8}$ &  $78.7\pm^{2.4}_{1.5}$ &  $82.9\pm^{0.3}_{2.7}$ &  $83.3\pm^{0.1}_{2.7}$ &  $83.6\pm^{-}_{0.7}$ &  $84.1\pm^{-}_{0.7}$ \\
 & BoW &   max &  LSTM &      - &  $67.1\pm^{10.1}_{15.7}$ &  $70.8\pm^{6.8}_{13.0}$ &  $79.3\pm^{2.8}_{4.5}$ &  $83.4\pm^{0.3}_{0.9}$ &  $83.9\pm^{0.1}_{0.9}$ &  $84.2\pm^{-}_{0.8}$ &  $85.0\pm^{-}_{0.2}$ \\
 &LSTM &   max &   Uni &      - &    $62.3\pm^{4.6}_{3.8}$ &   $68.5\pm^{4.7}_{4.3}$ &  $76.7\pm^{1.7}_{3.6}$ &  $80.4\pm^{0.2}_{3.2}$ &  $80.9\pm^{0.1}_{3.1}$ &  $81.1\pm^{-}_{0.9}$ &  $83.1\pm^{-}_{0.5}$ \\
 & LSTM &  last &   Uni &      - &    $65.0\pm^{8.0}_{4.4}$ &   $68.8\pm^{4.0}_{3.9}$ &  $74.1\pm^{2.0}_{1.4}$ &  $78.5\pm^{0.3}_{3.0}$ &  $79.1\pm^{0.1}_{3.1}$ &  $79.3\pm^{-}_{1.1}$ &  $81.6\pm^{-}_{0.5}$ \\
 & BoW &   max &   Uni &      - &    $59.9\pm^{7.2}_{3.7}$ &   $68.8\pm^{5.0}_{2.8}$ &  $77.3\pm^{1.2}_{0.9}$ &  $81.1\pm^{0.3}_{0.5}$ &  $81.5\pm^{0.1}_{0.5}$ &  $81.4\pm^{-}_{0.1}$ &  $83.3\pm^{-}_{0.4}$ \\
 & LSTM &   avg &  LSTM &      LM &    $63.6\pm^{7.4}_{5.4}$ &   $71.0\pm^{6.8}_{2.7}$ &  $81.0\pm^{1.6}_{2.3}$ &  $83.2\pm^{0.7}_{0.8}$ &  $83.8\pm^{0.1}_{0.8}$ &  $83.6\pm^{-}_{0.7}$ &  $84.4\pm^{-}_{0.5}$ \\
\bottomrule
\end{tabular}
\end{adjustbox}
\end{small}
\end{center}
\caption{Using \textit{BoW} encoders, \textit{Uni} decoders or \textit{PreLM} pretraining, the learned representations are more predictive of the labels (sentiment or topic) of the documents.}
\label{table_pooling}
\end{table*}

\begin{table}[t]
\begin{center}
\begin{adjustbox}{max width=0.34\textwidth}
\begin{tabular}{c|ccc}
\toprule
Dataset & F1(All) & F1(3) & Ratio \\
\midrule
AGNews & 89.0 & 71.9 & 0.808 \\
Amazon & 48.9 & 29.7 & 0.607 \\
Yahoo & 63.0 & 19.1 & 0.303 \\
Yelp & 96.5 & 82.4 & 0.854 \\
\bottomrule
\end{tabular}
\end{adjustbox}
\end{center}
\caption{Performance of bag-of-word classifiers when using all words as features versus only the first three words. Ratios of performance vary a lot across datasets.}
\label{table_bow_all_vs_three}
\end{table}

\subsection{Recurrent and \textit{BoW} encoders work around max-pooling}

\label{app_workaround}

It is counter-intuitive that \textit{BoW-max-LSTM} improves over \textit{LSTM-max-LSTM} (with or without \textit{PreAE}). Indeed, taking into account word order should allow the LSTM encoder to do better inference than the \textit{BoW} encoder, for example, by handling negation or parsing more complicated discourse structure \citep{pang_thumbs_2002}.

LSTM encoders are more powerful, but it can lead them to learn undesirable behaviors. We noticed that some components of the hidden states consistently reach their maximum values at fixed positions, regardless of the inputs (i.e., for some components $j{^*}$, $\argmax_i h_i^{j^*} \approx K$). These positions $K$ are often early positions in the sentence. For instance, with $\lambda=8$, $d=16$, \textit{LSTM-max-LSTM-PreAE} has 70 components out of 512 that are selected on 80\% of the documents on the same position on Yelp (68 on the first word, 2 on the second) and 78 on Amazon (57 on the first word, 21 on the second). In other words, some components of $r$ act like memory slots assigned to fixed positions in the sentence. This is probably achieved through counting mechanisms \citep{shi_why_2016,suzgun_lstm_2019}. The decoder is also an LSTM and can count, so it can also extract the relevant components at each position to retrieve the corresponding words. 

For \textit{BoW} encoders, it is less clear. It is possible that on some datasets, capitalized words could take especially high values on some components, in order to be consistently represented after max pooling. However, we have not explored the issue further. 

\section{Qualitative analysis}
\label{app_qualitative}

For our qualitative analysis, we take a look at the reconstruction samples (which were also used in Section \ref{section_text_gen}). We focus on the \textit{PreUni} models which lower memorization the most with the \textit{LSTM-max} and \textit{BoW-max} encoders, and compare them to the two best baselines. We use only one seed and one $z$ sample per model and per source sentence, but use two decoding strategies (beam search and greedy decoding).

In general, and for the reasons explained above, the rate $\lambda=8$ is chosen too small to recover exactly the source. Indeed, this rate is an order of magnitude less than the negative log-likelihood of \textit{LSTM-LM} baselines: above $80$ for all datasets except on Yelp where it is around $34.60$ nats (cf. Table \ref{table_datasets}). Since the NLL is an upper-bound on the entropy of the data, it gives a crude over estimate of the information content of the average document. On Yelp, where the NLL is much smaller (around $34.60$ nats), we hope to obtain good paraphrases for simple and frequent sentences. On the other datasets, we can not hope to reconstruct the sentences correctly but merely to control the generation by producing sentences which have the same labels as the source sentences. For this reason, we cherry-pick source sentences that look quite generic, because they are more probable and therefore, should be easy to reconstruct correctly.

Results are presented in Tables \ref{table_agnews_samples}, \ref{table_amazon_samples}, \ref{table_yahoo_samples} and \ref{table_yelp_samples}. Overall, we do observe less memorization of the first words and more correct sentiment or topic. Between our two models, on Amazon and Yelp (sentiment labels), it seems that \textit{LSTM-max-LSTM-PreUni} might perform better than \textit{BoW-max-LSTM-PreUni} because of its ability to handle negation, probably thanks to the recurrent encoder. It also seem more on topic on AGNews. On Yelp, it is able to paraphrase generic, small sentences. Therefore, we recommend this model as a future baseline.

We have already seen that it is very hard to classify sentences based on their first three words on the Amazon and Yahoo dataset, and that the baseline methods will learn representations that are not predictive of labels. However, the Yahoo dataset is especially challenging and our methods also struggle on it. We hypothesize that it is because only a few words per sentences are correlated with the labels. Indeed, there are many sentences of the form ``what do you think about \textit{X}?'' or ``what is the difference between \textit{X} and \textit{Y}?'' where only \textit{X} and \textit{Y} are correlated with the label and moreover often out-of-vocabulary. There might be no benefits for the model to diminish the reconstruction loss in priority on these words. By comparison, the other ``hard'' dataset (Amazon) is ``easier'', because the sentiment is often indicated by frequent adjectives like ``horrible'' or ``good'' and other frequent verbs.

\begin{table*}[t]
\begin{small}
\begin{center}
\begin{tabularx}{\textwidth}{l@{~}l@{~}|L}
\toprule
Model/``source" & Dec. & Sample \\
\midrule
source & &   Michael Owen heads England 's winner in the World Cup qualifier against Azerbaijan .  \\
\midrule
LSTM-last-LSTM-PreAE & beam &  Michael Owen will be sidelined for the rest of the season with a knee injury . \\
LSTM-last-LSTM-PreAE & greedy & Michael Owen has been charged with a rib injury and a new team for the first time in the last two weeks . \\
LSTM-max-LSTM-PreAE & beam &  American Bode Miller won the World Cup super @-@ G with a 6 @-@ 3 victory over the United States in the World Cup of Hockey . \\
LSTM-max-LSTM-PreAE & greedy & American Bode Miller won the World Cup of Hockey on Sunday , beating the United States by a record @-@ setting victory over the United States . \\
BoW-max-LSTM-PreUni & beam &  England coach Sven @-@ Goran Eriksson says he will not be able to win the World Cup qualifier against Wales . \\
BoW-max-LSTM-PreUni & greedy & England captain David Beckham has been named the England captain for the 2006 World Cup qualifiers against Wales .
\\
LSTM-max-LSTM-PreUni & beam &  England coach Sven @-@ Goran Eriksson says he will be fit for the World Cup qualifier against Wales next month . \\
LSTM-max-LSTM-PreUni & greedy & England captain David Beckham has been named the first World Cup qualifier in the World Cup qualifier against Wales . \\

\midrule
source & &   New Athlon 64 processors will compete with Intel 's Pentium 4 Extreme Edition .  \\
\midrule
LSTM-last-LSTM-PreAE & beam &  IBM ' s dual @-@ core Opteron processor will be available in the next three years . \\
LSTM-last-LSTM-PreAE & greedy & A new chipset for mobile phones will be available in the next three years . \\
LSTM-max-LSTM-PreAE & beam &  New version of Windows Server 2003 . \\
LSTM-max-LSTM-PreAE & greedy & New version of the Linux operating system is designed to integrate Linux and Linux . \\
BoW-max-LSTM-PreUni & beam &  Hewlett @-@ Packard Co . , the world 's largest computer maker , has unveiled a new version of its iPod digital music player , the company said . \\
BoW-max-LSTM-PreUni & greedy & Hewlett @-@ Packard Co . , the world 's largest maker of digital music player , on Tuesday unveiled a new version of its popular PlayStation 2 game console , which will be available in the next few years . \\
LSTM-max-LSTM-PreUni & beam &  Intel has unveiled a new version of its Pentium 4 Extreme Edition processor , which will be available for the first time . \\
LSTM-max-LSTM-PreUni & greedy & Intel has unveiled a new version of its Pentium M processor , which is designed to help the company 's new processor @-@ based processors . \\
\midrule
source & &   Nortel said it expects revenue for the third quarter to fall short of expectations .  \\
\midrule
LSTM-last-LSTM-PreAE & beam &  Coca @-@ Cola Co . \\
LSTM-last-LSTM-PreAE & greedy & research ) is expected to announce a new deal with the company to buy the company . \\
LSTM-max-LSTM-PreAE & beam &  Nortel Networks Corp. \\
LSTM-max-LSTM-PreAE & greedy & General Electric Co. said on Thursday it will buy the company for \$ 500 million in cash and stock . \\
BoW-max-LSTM-PreUni & beam &  Ford Motor Co . , the world 's largest maker of photographic film , said on Thursday it expects to cut its full @-@ year earnings forecast , citing strong demand for its flagship database products . \\
BoW-max-LSTM-PreUni & greedy & Coca @-@ Cola Co. on Thursday said third @-@ quarter earnings rose 29 percent , helped by strong sales of its soft drinks and business software . \\
LSTM-max-LSTM-PreUni & beam &  Nortel Networks Corp. , the world 's largest maker of equipment , said on Thursday that its third @-@ quarter profit rose 12 percent , helped by a rebound in the value of its assets . \\
LSTM-max-LSTM-PreUni & greedy & Shares of Nortel Networks Corp. fell nearly 8 percent on Thursday after the company said it expects its earnings for the third quarter , citing a decline in its third @-@ quarter earnings . \\
\bottomrule
\end{tabularx}
\end{center}
\end{small}
	\caption{Cherry-picked AGNews samples. Baselines are the first two models, our models are the two last. In the first example, the first baseline copies ``Michael Owen'' and complete with generic suffixes; the second baseline is about Hockey instead of soccer. Our baselines do not copy the beginning while correctly identifying the topic of England and the World Cup's qualifier. Similar comments can be made on the two other examples.}
\label{table_agnews_samples}
\end{table*}

\begin{table*}[t]
	\begin{small}
\begin{center}
	\begin{tabularx}{\textwidth}{l@{~}l@{~}|L}
\toprule
Model/``source" & Dec. & Sample \\
\midrule
source & &   I loved the book , but was a little bit UNK with the ending .  \\
\midrule
LSTM-last-LSTM-PreAE & beam &  I enjoyed the book , but the story line was not as good as the first one . \\
LSTM-last-LSTM-PreAE & greedy & I enjoyed the book and the story line . I was very disappointed in the book . \\
LSTM-max-LSTM-PreAE & beam &  I was very disappointed in the quality of the book , and the content of the book is very poor . \\
LSTM-max-LSTM-PreAE & greedy & I was disappointed in the quality of the book , but the book is not as good as the original . \\
BoW-max-LSTM-PreUni & beam &  I liked the story and the story line . It was a little slow at times but overall a good read . \\
BoW-max-LSTM-PreUni & greedy & I liked the story and the story line . It was a little slow but the ending was a little predictable . \\
LSTM-max-LSTM-PreUni & beam &  The book was a little slow , but the story line was good . I enjoyed it . \\
LSTM-max-LSTM-PreUni & greedy & The book was a little slow and the story line was very good . I was very disappointed . \\
\midrule
\midrule
source & &   This movie wasn 't as good as the original but I still enjoy watching it .  \\
\midrule
LSTM-last-LSTM-PreAE & beam &  This movie was a little slow at times , but it was a good movie . \\
LSTM-last-LSTM-PreAE & greedy & This movie was a little slow and the plot was not good . I would not recommend it to anyone . \\
LSTM-max-LSTM-PreAE & beam &  This movie was not as good as I thought it would be . I was very disappointed . \\
LSTM-max-LSTM-PreAE & greedy & The movie was not as good as the first one . I was disappointed in the quality of the movie . \\
BoW-max-LSTM-PreUni & beam &  Not as good as I thought it would be . I wouldn 't watch it again . \\
BoW-max-LSTM-PreUni & greedy & I didn 't like this movie . I thought it was going to be a good movie but I wouldn 't watch it again . \\
LSTM-max-LSTM-PreUni & beam &  I was expecting a little more from this movie . It was a little slow and boring . \\
LSTM-max-LSTM-PreUni & greedy & The movie was good , but the acting was not very good . I was expecting a little more from the movie . \\
\midrule
\midrule

source & &   This movie is horrible . The story , the acting , the directing . Just horrible .  \\
\midrule
LSTM-last-LSTM-PreAE & beam &  This is a great movie . I love it . It is a great family movie . \\
LSTM-last-LSTM-PreAE & greedy & This movie is great . It is a great movie and I love it . \\
LSTM-max-LSTM-PreAE & beam &  This is the worst movie I have ever seen . It was not worth the time to watch . \\
LSTM-max-LSTM-PreAE & greedy & This was a good movie . The acting was good , but the story line was not very good . \\
BoW-max-LSTM-PreUni & beam &  This movie is not worth the money . The acting is poor and the acting is poor . \\
BoW-max-LSTM-PreUni & greedy & The movie is very poor , the acting is poor . The acting is poor . \\
LSTM-max-LSTM-PreUni & beam &  This movie is a waste of time and money . It was a waste of time and money . \\
LSTM-max-LSTM-PreUni & greedy & This movie is a waste of time . It was a waste of time and money . \\
\midrule
\midrule

source & &   This book is very bad and does not give a real idea of the sport of UNK  \\
\midrule
LSTM-last-LSTM-PreAE & beam &  this is a great book for those who want to learn a little more about the history of the history of the history \\
LSTM-last-LSTM-PreAE & greedy & this book is a great book for the price , but the book is a little too short for my taste \\
LSTM-max-LSTM-PreAE & beam &  This book is a must have for anyone who is interested in the history of the Catholic Church \\
LSTM-max-LSTM-PreAE & greedy & This book is a must have for anyone who is interested in the field of the world of the New Testament \\
BoW-max-LSTM-PreUni & beam &  This is a good book , but it does not have a lot of information in it . \\
BoW-max-LSTM-PreUni & greedy & This book is a good book for the beginner , but it does not have a lot of information in it . \\
LSTM-max-LSTM-PreUni & beam &  This is a good book , but it does not have enough information on how to build your own . \\
LSTM-max-LSTM-PreUni & greedy & This is a good book , but the book is not a good book for the money . \\
\bottomrule
\end{tabularx}
\end{center}
\end{small}
	\caption{Cherry-picked Amazon samples. The first two examples are moderately positive reviews. LSTM-max-LSTM-PreUNI recreate roughly the same sentiment overall with different beginnings of sentences. On the third example, our models do copy the beginning of the source sentence but do not make mistakes on the sentiment as the baselines do. On the last, all models fail to capture the very negative sentiment, but our models at least moderate the positivity.}
\label{table_amazon_samples}
\end{table*}

\begin{table*}[t]
	\begin{small}
\begin{center}
	\begin{tabularx}{\textwidth}{l@{~}l@{~}|L}
\toprule
Model/``source" & Dec. & Sample \\
\midrule
source & &   What is your favorite book ?  \\
\midrule
LSTM-last-LSTM-PreAE & beam &  what is a good name for a girl ? \\
LSTM-last-LSTM-PreAE & greedy & what is a good name for a girl ? \\
LSTM-max-LSTM-PreAE & beam &  What is your favorite book ? \\
LSTM-max-LSTM-PreAE & greedy & What is your favorite book ? \\
BoW-max-LSTM-PreUni & beam &  What is your favorite sport ? \\
BoW-max-LSTM-PreUni & greedy & What is your favorite movie ? \\
LSTM-max-LSTM-PreUni & beam &  What is your favorite book ? \\
LSTM-max-LSTM-PreUni & greedy & What is your favorite Christmas movie ? \\
\midrule
\midrule

source & &   how can i get money ?  \\
\midrule
LSTM-last-LSTM-PreAE & beam &  i want to know ? \\
LSTM-last-LSTM-PreAE & greedy & i want to know ? \\
LSTM-max-LSTM-PreAE & beam &  how do i get money ? \\
LSTM-max-LSTM-PreAE & greedy & do you have any money ? \\
BoW-max-LSTM-PreUni & beam &  where can i go to watch free online ? \\
BoW-max-LSTM-PreUni & greedy & where can i get free online games ? \\
LSTM-max-LSTM-PreUni & beam &  where can i get a job ? \\
LSTM-max-LSTM-PreUni & greedy & how do i get a job in USA ? \\
\midrule
\midrule
source & &   What countries have nuclear weapons ?  \\
\midrule
LSTM-last-LSTM-PreAE & beam &  What are the pros and cons ? \\
LSTM-last-LSTM-PreAE & greedy & What are the pros and cons of smoking ? \\
LSTM-max-LSTM-PreAE & beam &  which countries in the world ? \\
LSTM-max-LSTM-PreAE & greedy & which country has the highest world cup ? \\
BoW-max-LSTM-PreUni & beam &  how many countries are there in the world ? \\
BoW-max-LSTM-PreUni & greedy & what are the countries that will be able to be the most effective government in the world ? \\
LSTM-max-LSTM-PreUni & beam &  Why are the colors of the Earth ? \\
LSTM-max-LSTM-PreUni & greedy & what are the three different countries in the U.S. ? \\
\midrule
\midrule

source & &   how to print all webpage content ?  \\
\midrule
LSTM-last-LSTM-PreAE & beam &  how to create a website ? \\
LSTM-last-LSTM-PreAE & greedy & how to find a website ? \\
LSTM-max-LSTM-PreAE & beam &  how can i learn english language ? \\
LSTM-max-LSTM-PreAE & greedy & how can i watch free online online ? \\
BoW-max-LSTM-PreUni & beam &  how do you get a free copy of the internet ? \\
BoW-max-LSTM-PreUni & greedy & how to get the free internet explorer ? \\
LSTM-max-LSTM-PreUni & beam &  how do i get a copy of my computer in the internet ? \\
LSTM-max-LSTM-PreUni & greedy & how do i get the power to open a computer in the internet ? \\

\bottomrule
\end{tabularx}
\end{center}
\end{small}
	\caption{Cherry-picked Yahoo samples. There isn't a model that clearly stands out, but we can rule out LSTM-last-LSTM-PreAE. This dataset is more difficult (see main text).}
\label{table_yahoo_samples}
\end{table*}

\begin{table*}[t]
	\begin{small}
\begin{center}
	\begin{tabularx}{\textwidth}{l@{~}l@{~}|L}
\toprule
Model/``source" & Dec. & Sample \\
\midrule
source & &   amazing place .  \\
\midrule
LSTM-last-LSTM-PreAE & beam &  amazing customer service . \\
LSTM-last-LSTM-PreAE & greedy & amazing customer service . \\
LSTM-max-LSTM-PreAE & beam &  amazing food . \\
LSTM-max-LSTM-PreAE & greedy & amazing food . \\
BoW-max-LSTM-PreUni & beam &  this place is amazing . \\
BoW-max-LSTM-PreUni & greedy & this place is amazing . \\
LSTM-max-LSTM-PreUni & beam &  this place is amazing . \\
LSTM-max-LSTM-PreUni & greedy & this place is amazing . \\
\midrule
\midrule

source & &   definitely going back soon !  \\
\midrule
LSTM-last-LSTM-PreAE & beam &  definitely coming back ! \\
LSTM-last-LSTM-PreAE & greedy & definitely recommend to anyone ! \\
LSTM-max-LSTM-PreAE & beam &  definitely coming back again ! \\
LSTM-max-LSTM-PreAE & greedy & highly recommend them to anyone ! \\
BoW-max-LSTM-PreUni & beam &  i will definitely be back ! \\
BoW-max-LSTM-PreUni & greedy & i will definitely be back ! \\
LSTM-max-LSTM-PreUni & beam &  i will be back ! \\
LSTM-max-LSTM-PreUni & greedy & i will be back ! \\
\midrule
\midrule
source & &   not worth the risk .  \\
\midrule
LSTM-last-LSTM-PreAE & beam &  not the best . \\
LSTM-last-LSTM-PreAE & greedy & not the best . \\
LSTM-max-LSTM-PreAE & beam &  not worth the money . \\
LSTM-max-LSTM-PreAE & greedy & not worth the money . \\
BoW-max-LSTM-PreUni & beam &  worth the wait . \\
BoW-max-LSTM-PreUni & greedy & it was worth the wait . \\
LSTM-max-LSTM-PreUni & beam &  not worth the hassle . \\
LSTM-max-LSTM-PreUni & greedy & it 's not worth the money . \\
\midrule
\midrule

source & &   overall , a huge disappointment .  \\
\midrule
LSTM-last-LSTM-PreAE & beam &  pizza was good too . \\
LSTM-last-LSTM-PreAE & greedy & pizza was good too . \\
LSTM-max-LSTM-PreAE & beam &  ok , nothing special . \\
LSTM-max-LSTM-PreAE & greedy & nothing special , but the food was bland . \\
BoW-max-LSTM-PreUni & beam &  wow . \\
BoW-max-LSTM-PreUni & greedy & great experience . \\
LSTM-max-LSTM-PreUni & beam &  what a disappointment . \\
LSTM-max-LSTM-PreUni & greedy & what a disappointment . \\
\bottomrule
\end{tabularx}
\end{center}
\end{small}
	\caption{Cherry-picked Yelp samples. On small and typical sentences, our last variant LSTM-max-LSTM-PreUni can produce paraphrases. On the other hand, BoW-max-LSTM-PreUni fails on the two negative examples, probably because it lacks the ability to deal with negation. The baseline models also fail to capture the sentiment on the last example, and copy the beginning on the first three examples.}
\label{table_yelp_samples}
\end{table*}

\end{document}